\documentclass[10pt,twocolumn,letterpaper]{article}

\usepackage[accsupp]{axessibility}  
\usepackage{iccv}
\usepackage{times}
\usepackage{epsfig}
\usepackage{graphicx}
\usepackage{amsmath}
\usepackage{amssymb}
\usepackage{comment}
\usepackage[table]{xcolor}
\usepackage{subfigure}
\usepackage{epstopdf}
\usepackage[]{booktabs}
\usepackage{multirow}
\usepackage{siunitx}

\usepackage{xprintlen}

\newcommand{\beforefigcaption}{\vspace{-5mm}}
\newcommand{\afterfigcaption}{\vspace{-5mm}}
\newcommand{\beforetab}{\vspace{-1mm}}
\newcommand{\aftertab}{\vspace{-4.5mm}}
\newcommand{\beforesection}{\vspace{-1.5mm}} 
\newcommand{\aftersection}{\vspace{-1.5mm}}
\newcommand{\beforesubsection}{\vspace{-1.5mm}}
\newcommand{\aftersubsection}{\vspace{-1.5mm}}

\usepackage{etoolbox}
\newcommand{\zerodisplayskips}{%
  \setlength{\abovedisplayskip}{3pt}%
  \setlength{\belowdisplayskip}{3pt}%
  \setlength{\abovedisplayshortskip}{0pt}%
  \setlength{\belowdisplayshortskip}{0pt}}
\appto{\normalsize}{\zerodisplayskips}
\appto{\small}{\zerodisplayskips}
\appto{\footnotesize}{\zerodisplayskips}

\makeatletter
\newcommand{\thickhline}{%
    \noalign {\ifnum 0=`}\fi \hrule height 1pt
    \futurelet \reserved@a \@xhline
}
\makeatother

\usepackage[pagebackref=true,breaklinks=true,colorlinks,bookmarks=false]{hyperref}

\iccvfinalcopy 

\def\iccvPaperID{5420} 

\ificcvfinal\fi

\begin{document}

\title{
ARCH++: Animation-Ready Clothed Human Reconstruction Revisited
}

\author{Tong He$^{1,2}$\thanks{\,This work was done as part of Tong He's internship at Facebook, Sausalito, CA, USA. The corresponding author is Yuanlu Xu.},\ \ Yuanlu Xu$^{1*}$,\ \ Shunsuke Saito$^{1}$,\ \ Stefano Soatto$^{2}$,\ \ Tony Tung$^{1}$\\
$^{1}$Facebook Reality Labs Research, USA\quad\quad $^{2}$University of California, Los Angeles, USA\\
{\tt \small \{simpleig,soatto\}@cs.ucla.edu, \{merayxu,shunsuke.saito16\}@gmail.com, tony.tung@fb.com}
}

\maketitle
\ificcvfinal\thispagestyle{empty}\fi

\begin{abstract}
\vspace{-2mm}

We present ARCH++, an image-based method to reconstruct 3D avatars with arbitrary clothing styles. Our reconstructed avatars are animation-ready and highly realistic, in both the visible regions from input views and the unseen regions. While prior work shows great promise of reconstructing animatable clothed humans with various topologies, we observe that there exist fundamental limitations resulting in sub-optimal reconstruction quality. In this paper, we revisit the major steps of image-based avatar reconstruction and address the limitations with ARCH++. First, we introduce an end-to-end point based geometry encoder to better describe the semantics of the underlying 3D human body, in replacement of previous hand-crafted features. Second, in order to address the occupancy ambiguity caused by topological changes of clothed humans in the canonical pose, we propose a co-supervising framework with cross-space consistency to jointly estimate the occupancy in both the posed and canonical spaces. Last, we use image-to-image translation networks to further refine detailed geometry and texture on the reconstructed surface, which improves the fidelity and consistency across arbitrary viewpoints. In the experiments, we demonstrate improvements over the state of the art on both public benchmarks and user studies in reconstruction quality and realism. Project page: \url{https://tonghehehe.com/archpp}.

\vspace{-1mm}
\end{abstract}

\vspace{-2mm}
\beforesection
\section{Introduction} \label{sec:intro}
\aftersection

\begin{figure}[ptb]
\centering
\includegraphics[width=\linewidth]{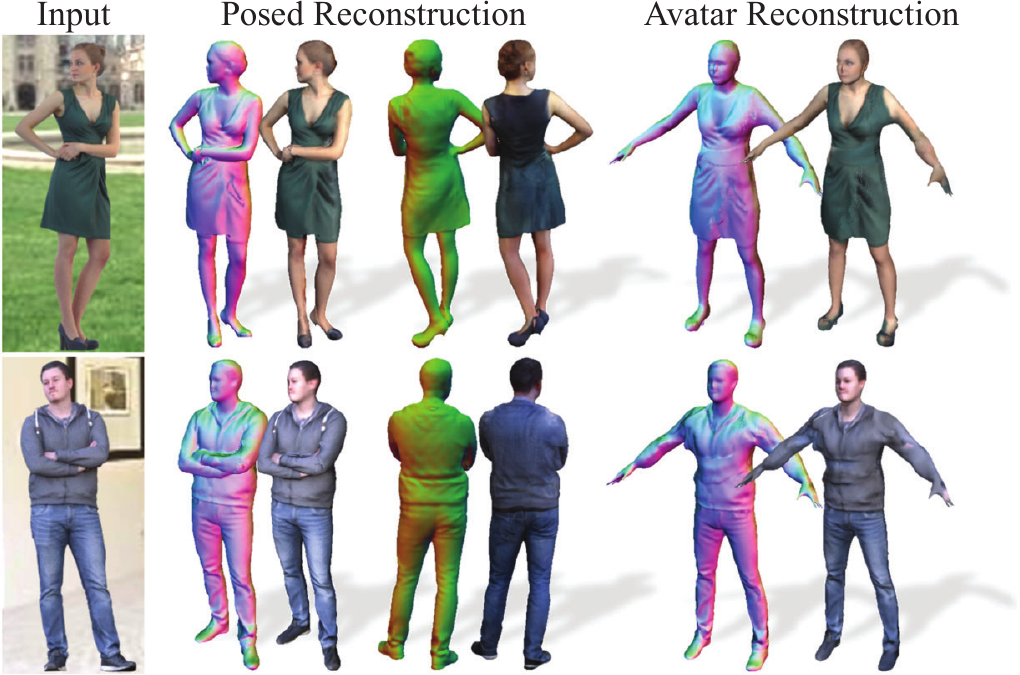}
\beforefigcaption
\caption{Given an image of a subject in arbitrary pose (left), our method could generate photorealistic avatars in both the posed input space (middle) as well as auto-rigged canonical space (right).}
\afterfigcaption
\label{fig:intro}
\end{figure}

Digital humans have become an increasingly important building block for numerous AR/VR applications, such as video games, social telepresence~\cite{orts2016holoportation, lombardi2018deep} and virtual try-on. Towards truly immersive experiences, it is crucial for these avatars to obtain higher level of realism that goes beyond the uncanny valley~\cite{mori2012uncanny}. Building a photorealistic avatar involves many manual works by artists or expensive capture systems under controlled environments~\cite{collet2015high, guo2019relightables, mathias21unoc}, limiting access and increasing cost. Therefore, it is vital to revolutionize reconstruction techniques with minimal prerequisite (\eg, a selfie) for future digital human applications.

Recent human models reconstructed from a single image combine category-specific data prior with image observations~\cite{andrei20nflow,joo2020eft,xu21posegrammar}. Among which, template-based approaches~\cite{kanazawa2018hmr, SPINICCV19, DenseRaCICCV19, Tex2ShapeICCV19, bhatnagar2019mgn} nevertheless suffer from lack of fidelity and difficulty supporting clothing variations; 
while non-parametric reconstruction methods~\cite{PIFuICCV19, DeepHumanICCV19, saito2020pifuhd, he2020geopifu}, \eg, using implicit surface functions, do not provide intuitive ways to animate the reconstructed avatar despite impressive fidelity. 
In the recent work ARCH~\cite{huang2020arch}, the authors propose reconstructing non-parametric human model using pixel-aligned implicit functions~\cite{PIFuICCV19} in a canonical space, where all reconstructed avatars are transformed to a common pose. To do so, a parametric human body model is exploited to determine the transformations. By transferring skinning weights, which encode how much each vertex is influenced by the transformation of each body joint, from the underling body model, the reconstruction results are ready to animate. However, we observe that the advantages of a parametric body model and pixel-aligned implicit functions are not fully exploited. 

In this paper we introduce ARCH++, which revisits the major steps of animatable avatar reconstruction from images and addresses the limitations in the formulation and representation of the prior work.
First, current implicit function based methods mainly use hand-crafted features as the 3D space representation, which suffers from depth ambiguity and lacks human body semantic information. To address this, we propose an end-to-end geometry encoder based on PointNet++~\cite{qi2017pointnet,qi2017pointnet++}, which expressively describes the underlying 3D human body. 
Second, we find the unposing process to obtain the canonical space supervision causes topology change (\eg, removing self-intersecting regions) and consequently the articulated reconstruction fails to obtain the same level of accuracy in the original posed space.
Therefore, we present a co-supervising framework where occupancy is jointly predicted in both the posed and canonical spaces, with additional constraints on the cross-space consistency. This way, we benefit from both: supervision in the posed space allows the prediction to retain all the details of the original scans; while canonical space reconstruction can ensure the completeness of a reconstructed avatar. 
Last, image-based avatar reconstruction often suffers from degraded geometry and texture in the occluded regions. To make the problem more tractable, we first infer surface normals and texture of the occluded regions in the image domain using image translation networks, and then refine the reconstructed surface with a moulding-inpainting scheme.

In the experiments, we evaluate ARCH++ on photorealistically rendered synthetic images as well as in-the-wild images, outperforming prior works based on implicit functions and other design choices on public benchmarks.

\begin{figure*}[ptb]
\vspace{-2mm}
\centering
\includegraphics[width=\textwidth]{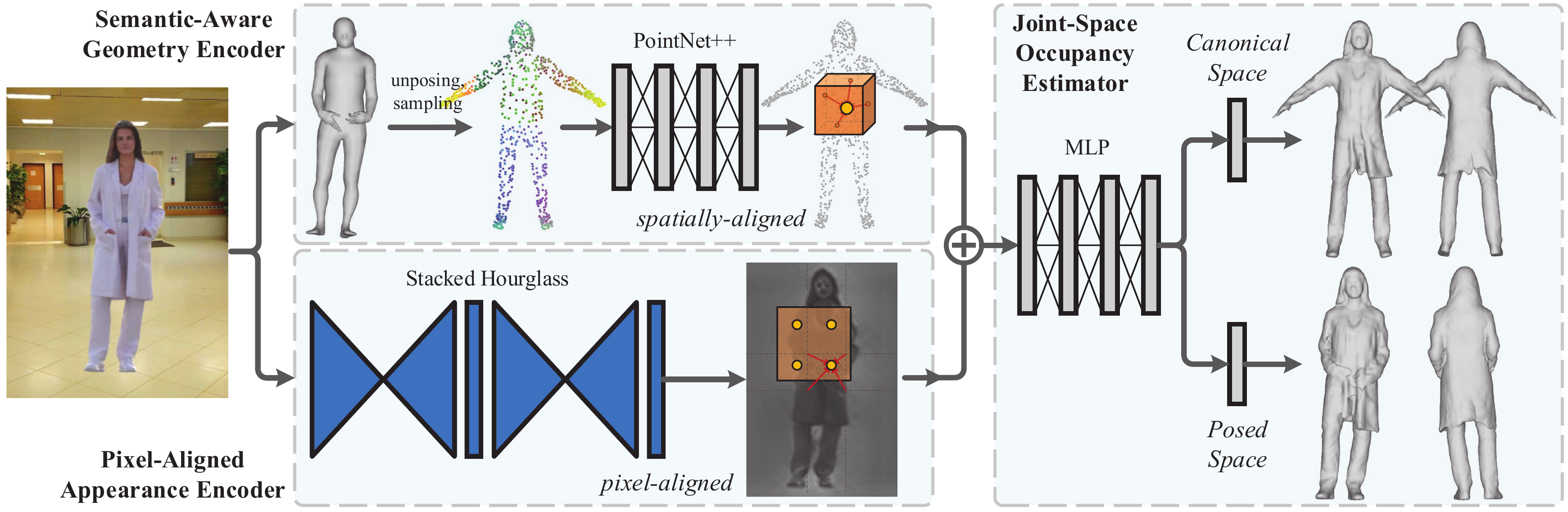}
\beforefigcaption
\vspace{0.5mm}
\caption{\emph{Overview of the initial joint-space implicit surface reconstruction}. This procedure includes three components: i) semantic-aware geometry encoder, ii) pixel-aligned appearance encoder and iii) joint-space occupancy estimator. See text for detailed explanation.
}
\afterfigcaption
\label{fig:framework}
\end{figure*}

The contributions of ARCH++ include: 
1) a point-based geometry encoder for implicit functions to directly extract human shape and pose priors, which is efficient and free from quantization errors;
2) we are the first to point out and study the fundamental issue of determining target occupancy space: posed-space fidelity vs. canonical-space completeness. Albeit ignored before, we outline the pros and cons of different spaces, and propose a co-supervising framework of occupancy fields in joint spaces;
3) we discover image-based surface attribute estimation could address the open problem of view-inconsistent reconstruction quality. Our moulding-inpainting surface refinement strategy generates 360$^{\circ}$ photorealistic 3D avatars.
4) our method demonstrates enhanced performance on the brand new task of image-based animatable avatar reconstruction.

\beforesection\vspace{1mm}
\section{Related Work} \label{sec:related}
\aftersection\vspace{1mm}

\textbf{Template-based reconstruction} utilizes parametric human body models, \eg, SCAPE~\cite{angelov05} and SMPL~\cite{loper2015smpl} to provide strong prior on body shape and pose to address ill-posed problems including body estimation under clothing~\cite{yangClothMocapECCV16, zhang-CVPR17} and image-based human shape reconstruction~\cite{bogo2016keep, lassner2017unite, kanazawa2018hmr, HoloPoseCVPR19, CMRCVPR19, xiang2019monocular, SPINICCV19, DenseRaCICCV19}. While these works primarily focus on underling body shapes without clothing, the template-based representations are later extended to modeling clothed humans with displacements from the minimal body~\cite{ClothCapTOG17}, or external clothing templates~\cite{MulGarmentNetICCV19}, from 3D scans~\cite{yang2018analyzing, ClothCapTOG17}, videos~\cite{Video3DPeopleCVPR18, marc2020deepcap}, and a single image~\cite{Video3DClothPeopleCVPR19, MulGarmentNetICCV19, jiang2020bcnet}. As these approaches build clothing shapes on a body template mesh, the reconstructed models can be easily driven by pose parameters of the parametric body model. To address the lack of details with limited mesh resolutions, recent works propose to utilize 2D UV maps~\cite{laehnerECCV18,Tex2ShapeICCV19}. However, as a clothing topology can significantly deviate from the underling body mesh and its variation is immense, these template-based solutions fail to capture clothing variations in the real world.

\textbf{Non-parametric capture} is widely used to capture highly detailed 3D shapes with an arbitrary topology from multi-view systems under controlled environments~\cite{MVVisualHullTOG00, MVDetailPoseShapeCVPR07, MVArticulatedMeshAnimeTOG08, tung08, MVSkelSurfMocapCVPR09, MVPhotoStereoTOG09, MVStereopsisTOG10, MVShadeMocapECCV12, tung09, tung12}. Recent advances of deep learning further push the envelope by supporting sparse view inputs~\cite{gilbert2018volumetric, huang2018deep}, and even monocular input~\cite{li2020monocular}. For single-view clothed human reconstruction, direct regression methods demonstrate promising results, supporting various clothing types with a wide range of shape representations including voxels~\cite{varol18_bodynet, VolumeRegECCVW2018}, two-way depth maps~\cite{gabeur2019moulding, smith2019facsimile}, visual hull~\cite{SiCloPeCVPR19}, and implicit functions~\cite{PIFuICCV19, saito2020pifuhd, he2020geopifu}. In particular, pixel-aligned implicit functions (PIFu)~\cite{PIFuICCV19} and its follow-up works~\cite{saito2020pifuhd, he2020geopifu} demonstrate impressive reconstruction results by leveraging neural implicit functions~\cite{OccupancyNetCVPR19, chen2019implicit_decoder, Park_2019_CVPR} and fully convolutional image features. Unfortunately, despite its high-fidelity results, non-parametric reconstructions are not animation-ready due to missing body part separation and articulation. Recently, IF-Net~\cite{chibane20ifnet} exploits partial point cloud inputs and learns implicit functions using latent voxel features. Compared with image-based avatar reconstruction, completion from points can leverage directly provided strong shape and pose cues, and thus skip learning them from complex images.

\textbf{Hybrid approaches} combine template-based and non-parametric methods and allow us to leverage the best of both worlds, namely structural prior and support of arbitrary topology. Recent work~\cite{bharat20ipnet} shows that using SMPL model as guidance significantly improves robustness of non-rigid fusion from RGB-D inputs. For single-view human reconstruction, Zheng et al. first introduce a hybrid approach of a template-model (SMPL) and a non-parametric shape representation (voxel~\cite{DeepHumanICCV19} and implicit surface~\cite{Zerong2020PaMIR}). These approaches, however, choose an input view space for shape modeling with reconstructed body parts potentially glued together, making the reconstruction difficult to animate as in the aforementioned non-parametric methods. The most relevant work to ours is ARCH~\cite{huang2020arch}, where the reconstructed clothed humans are ready for animation as pixel-aligned implicit functions are modeled in an unposed canonical space. However, such framework fundamentally leads to sub-optimal reconstruction quality. We achieve significant improvement on accuracy and photorealism by addressing the hand-crafted spatial encoding for implicit functions, the lack of supervision in the original posed space, and the limited fidelity of occluded regions.

\beforesection
\section{Proposed Methods} \label{sec:method}
\aftersection

Our proposed framework, ARCH++, uses a coarse-to-fine scheme,
\ie, initial reconstruction by learning joint-space implicit surface functions (see Fig.~\ref{fig:framework}), and then mesh refinement in both spaces (see Fig.~\ref{fig:refine}).

\beforesubsection
\subsection{Joint-Space Implicit Surface Reconstruction} \label{sec:model}
\aftersubsection

\textbf{Semantic-Aware Geometry Encoder}. The spatial feature representation of a query point is critical for deep implicit function. While the pixel-aligned appearance feature via Stack Hourglass Network~\cite{newell2016stacked} has already demonstrated its effectiveness in detailed clothed human reconstruction by prior works~\cite{PIFuICCV19,saito2020pifuhd,huang2020arch,he2020geopifu}, an effective design of point-wise spatial encoding has not yet been well studied. The extracted geometry features should be informed of the semantics of the underlying 3D human body, which provide strong priors to regularize the overall dressed people shape. 

The spatial encoding methods used previously include hand-crafted features (\eg, RBF~\cite{huang2020arch}) and latent voxel features~\cite{chibane20ifnet,he2020geopifu,Zerong2020PaMIR}. The former is constructed based on Euclidean distances between a query point and the body joints, ignoring the shapes. The voxel-based features capture both shape and pose priors of a parametric body mesh. Compared with the hand-crafted features, the end-to-end learned voxel features are better informed of the underlying body structures but often constrained by GPU memory sizes and suffer from quantization errors due to low spatial resolution. To effectively encode the shape and pose priors without losing any precision, we propose a novel semantic-aware geometry encoder that extracts point-wise spatial encodings. Essentially a parametric body mesh can be sampled into a point cloud and fed into PointNet++~\cite{qi2017pointnet,qi2017pointnet++} to learn point-based spatial features, which have several advantages over both hand-crafted RBF features and voxel-based ones. Our method encodes both shape and pose priors from parametric shapes without computation overhead and quantization errors caused by the mesh voxelization process. Additional detailed statistical comparisons on points v.s. voxels in representing 3D shapes are reported in~\cite{gabeur2019moulding}.

Given a parametric body mesh estimated and deformed by~\cite{DenseRaCICCV19, huang2020arch}, we use a PointNet++~\cite{qi2017pointnet,qi2017pointnet++} based semantic-aware geometry encoder to learn the underlying 3D human body prior.
We sample $N_0$ (\eg, 7324) points from the body mesh surfaces and feed them into
the geometry encoder
for spatial feature learning, that is,
\begin{equation} \small \begin{aligned}
f_{pn}:\{x_0^i\}_{i=1}^{N_0} \mapsto \{x_1^j,h_1^j\}_{j=1}^{N_1},\{x_2^k,h_2^k\}_{k=1}^{N_2},\{x_3^l,h_3^l\}_{l=1}^{N_3},
\end{aligned} \end{equation}
where $x_0^i \in \mathbb{R}^{3}$ is a point sampled from the parametric body mesh. The PointNet++ based encoder utilizes fully-connected layers and neighborhood Max-Pooling to extract semantic-aware geometry features $h \in \mathbb{R}^{32}$ of a point. It also applies Furthest Point Sampling to progressively down sample the points $N_1=2048, N_2=512, N_3=128$ to extract latent features with increasing receptive fields. For example $\{x_1^j\}$ is a down sampled point set with size $N_1$, and $h_1^j \in \mathbb{R}^{32}$ is the learned feature \wrt each point.

As illustrated in Fig.~\ref{fig:framework}, for any query point $p_a \in \mathbb{R}^{3}$ in the canonical space we obtain its point-wise spatial encoding $f_g \in \mathbb{R}^{96}$
via inverse $L2$-norm kernel based feature interpolation, followed by query coordinates concatenated Multi-layer Perceptrons (MLP).
Particularly, we extract these features from different point set densities-$j,k,l$ to construct concatenated features $f_g = (f_g^j \oplus f_g^k \oplus f_g^l)$ that are informed of multi-scale structures. For example, $f_g^j \in \mathbb{R}^{32}$ is defined as:
\begin{equation} \small \begin{aligned}
f_g^j(p_a, \{x_1^j,h_1^j\}) &= \text{MLP}(p_a \oplus \sum_{m} \frac{\left \| p_a - x_1^m \right \|^{-2}}{S(p_a, \{x_1^j,h_1^j\})}h_1^m),\\
S(p_a, \{x_1^j,h_1^j\}) &= \sum_{m} \left \| p_a - x_1^m \right \|^{-2},
\end{aligned} \end{equation}
where the index $m$ is determined by finding the K nearest neighbors among the point set $\{x_1^j\}$ \textit{w.r.t.} the query point. Empirically we found setting $K=3$ obtains fair performance. The features extracted at other point set densities $f_g^k, f_g^l \in \mathbb{R}^{32}$ are obtained similarly leveraging $\{x_2^k,h_2^k\}$ and $\{x_3^l,h_3^l\}$, respectively.

\textbf{Pixel-Aligned Appearance Encoder}. We share the same architecture design as~\cite{PIFuICCV19,saito2020pifuhd,huang2020arch,he2020geopifu} to map an input image $I \in \mathbb{R}^{512\times512\times3}$ into the latent feature maps $\psi_{\mu}(I) \in \mathbb{R}^{128\times128\times256}$ via a Stacked Hourglass Network~\cite{newell2016stacked} with weights $\mu$. To obtain appearance encoding $f_a \in \mathbb{R}^{256}$ of any query point $p_b \in \mathbb{R}^{3}$ in the posed space, we project it back to the image plane based on a camera model of weak perspective projection, and bilinearly interpolate the latent image features:
\begin{equation} \small \begin{aligned}
f_a(p_b,I) = \mathcal{B}(\psi_{\mu}(I), \pi(p_b)),
\end{aligned} \end{equation}
where $\mathcal{B}(\cdot)$ indicates the differentiable bilinear sampling operation, and $\pi(\cdot)$ means weak perspective camera projection from the query point $p_b$ to the image plane of $I$.

\textbf{Joint-Space Occupancy Estimator}. While most non-parametric and hybrid methods use the posed space as the learning and inference target space, ARCH instead reconstructs the clothed human mesh directly in a canonical space where humans are in a normalized A-shape pose. Different choices of the target space have pros and cons.
The posed space is naturally aligned with the input pixel evidence and therefore the reconstructions have high data fidelity leveraging the direct image feature correspondences. Thus, many works choose to reconstruct a clothed human mesh in its original posed space (\eg, PIFu(HD)~\cite{PIFuICCV19,saito2020pifuhd}, Geo-PIFu~\cite{he2020geopifu}, PaMIR~\cite{Zerong2020PaMIR}). However, in many situations the human can demonstrate complex poses with self-intersection (\eg, hands in the pocket, crossed arms) and cause a "glued" mesh that is difficult to articulate. Meanwhile, canonical pose reconstruction offers us a rigged mesh that is animation ready 
(via its registered A-shape parametric mesh~\cite{huang2020arch}).
The problem of using the canonical space as the target space is that when we warp the mesh into its posed space there could be artifacts like intersecting surfaces and distorted body parts (see Fig.~\ref{fig:ablation_space}). Thus, the reconstruction fidelity of the warping obtained canonical-to-posed space mesh will degenerate. To maintain both input image fidelity and reconstruction surface completeness, we propose to learn the joint-space occupancy distributions.

We use a joint-space defined occupancy map $O$ to implicitly represent the 3D clothed human under both its original posed space and a rigged canonical space:
\begin{equation}\small
    O = \{ (p_a, p_b, o_a, o_b):\; p_a,p_b \in \mathbb{R}^3,\ -1 \leq o_a, o_b \leq 1\},
\end{equation}
where $o_a,o_b$ denote the occupancy for points $p_a$ and $p_b$. A point in the posed space is $p_b$ and its mapped counterpart in the canonical space is $p_a = \text{SemDF}(p_b)$. The semantic deformation mapping (SemDF) between the original posed and the rigged canonical spaces is enabled by nearest neighbor-based skinning weights matching between $p_b$ and the estimated underlying parametric body mesh~\cite{huang2020arch}.

To enable mesh reconstruction in joint spaces, we use both point-wise spatial features $f_g \in \mathbb{R}^{96}$ that are informed of semantic full-body structures, and pixel-aligned features $f_a \in \mathbb{R}^{256}$ that encode human front-view appearances:
\begin{equation} \small \begin{aligned}
o_a = \mathcal{F}_{\theta}(f_g \oplus f_a),\ o_b = \mathcal{F}_{\beta}(f_g \oplus f_a),
\end{aligned} \end{equation}
where $\theta, \beta$ are network weights of the MLP-based deep implicit surface functions.
To reconstruct avatars from the dense occupancy estimations in two spaces, we use Marching Cube~\cite{MarchCube87} to extract the isosurface
at $o_a = \tau$ and $o_b = \tau$ (\ie, $\tau=0$), respectively.

The network outputs $o_a,o_b$ are supervised by the ground truth joint-space occupancy $\hat{o}_a,\hat{o}_b$, depending on whether a posed space query point $p_b$ and its corresponding canonical space point $p_a$ are inside the clothed human meshes or not. Though $p_a,p_b$ are a pair of mapped points their ground-truth occupancy values are not the same in all cases.
For example, a point outside and close to the hand of a parametric body could has $\hat{o}_b > 0$ and $\hat{o}_a < 0$ if the original mesh in posed space has self-contact (\eg, hands in the pocket).
Namely, the SemDF defines a dense correspondence mapping between the two spaces but their occupancy values are not necessarily the same. Therefore, naively learning the distribution in one space and then warping the reconstruction into another pose can cause mesh artifacts (see Fig.~\ref{fig:ablation_space}). This motivates us to model two space occupancy distributions jointly in order to maintain both canonical space mesh completeness and posed space reconstruction fidelity.

\beforesubsection
\subsection{Mesh Refinement} \label{sec:mesh_refine}
\aftersubsection

\begin{figure}[ptb]
\vspace{-2mm}
\centering
\includegraphics[width=\linewidth]{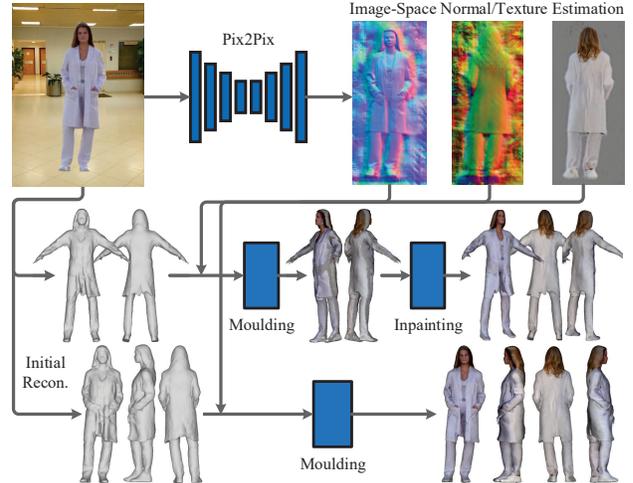}
\beforefigcaption
\caption{\emph{Overview of the mesh refinement steps}. Our approach refines the initially estimated joint-space meshes from Fig.~\ref{fig:framework} using estimated normals and textures.}
\afterfigcaption
\label{fig:refine}
\end{figure}

We further refine the reconstructed meshes in joint spaces by adding geometric surface details and photorealistic textures. As illustrated in Fig.~\ref{fig:refine}, we propose a moulding-inpainting scheme to utilize the front and back side normals and textures estimated in the image space. This is based on the observation that direct learning and inference of dense normal/color fields using deep implicit functions as~\cite{huang2020arch} usually leads to over-smooth blur patterns and block artifacts (see Fig.~\ref{fig:result_table}).
In contrast, image space estimation of normal and texture maps produces sharp results with fine-scale details, and is robust to human pose and shape changes. These benefits are from well-designed 2D convolutional deep networks (\eg, Pix2Pix~\cite{isola2017image,wang2018high}) and advanced (adversarial) image generation training schemes like GAN, with perceptual losses. The image-space estimated normal (and texture) maps could be used in two different ways. They can be used as either direct inputs into the Stack Hourglass as additional channels of the single-view image, or moulding-based front and back side mesh refinement sampling sources. In the experiments, we conduct ablation studies on these two schemes (\ie, early direct input, late surface refinement) and demonstrate that our moulding-based refinement is better at maintaining fine-scale surface details across different views (see Fig.~\ref{fig:ablation_normal}).

\textbf{Posed Space}. For the clothed human mesh obtained by Marching Cube in the original posed space, we conduct visibility tracing to determine if a vertex $V \in \mathbb{R}^3$ should be projected onto the front or the back side to bilinearly sample the normal/texture maps. Essentially, this is a moulding-based mesh refinement process for surface details and textures enhancement. We first conduct normal refinement. Note that for vertices whose unrefined normals $\mathbf{n} \in \mathbb{R}^3$ are near parallel (\ie, within $\varepsilon$ degrees) to the input image plane,
we project them onto both the front and the back side normal maps $I_{n}^f, I_{n}^b \in \mathbb{R}^{512\times512\times3}$. We could then compute the refined surface normals $\mathbf{n}' \in \mathbb{R}^3$ via a linear blend fusion:
\begin{equation} \small \begin{aligned}
\mathbf{n}' &= \chi (1-\alpha')\,\mathcal{B}(I_{n}^f,\pi(V)) + \chi (\alpha')\,\mathcal{B}(I_{n}^b,\pi(V)),\\
\alpha' &= (90^{\circ} + \varepsilon - \alpha)/(2\varepsilon),\\
\end{aligned} \end{equation}
where $\alpha$ is the angle between the unrefined normal and the forward camera raycast, and ${\alpha}'$ is the normalized value of $\alpha$. Again, $\mathcal{B}(\cdot)$ indicates the bilinear sampling operation. The indicator function $\chi(\cdot)$ determines the blending weights of sampled normals from the front and the back sides:
\begin{equation} \small
\chi({\alpha}') = \min(\max({\alpha}',\, 0),\, 1)
\end{equation}

This simple yet effective fusion scheme creates a normal-refined mesh with negligible blending boundary artifacts. With the refined surface normals we can further apply Poisson Surface Reconstruction~\cite{2013possion_recon} to update the mesh topology but in practice we find this unnecessary since the moulding-refined avatar can already satisfy various AR/VR and novel-view rendering applications. This bump rendering idea is also used in DeepHuman~\cite{DeepHumanICCV19} but they only refine meshes using the front views.
We further conduct the texture refinement in a similar manner but use the refined normals to help determine the linear blending weights of boundary vertices. Our moulding-based front/back normal and texture refinement method yields clothed human meshes that look photorealistic at different viewpoints with full-body surface details (\eg, clothes wrinkles, hairs).

\textbf{Canonical Space}. The reconstructed canonical space avatar is rigged and thus can be warped back to its posed space and then refined via the same pipeline described above. However, a unique challenge for canonical avatar refinement is that mesh reconstructions in this space might contain unseen surfaces under the posed space.
For example, in the third row of Fig.~\ref{fig:result_table}, the folded arm is in contact with the chest in the posed space but unfolded in the canonical space. Therefore, we do not have direct normal/texture correspondences of the chest regions of the canonical mesh. 
To address this problem, we render the front and the back side images of the canonical mesh with incomplete normal and texture, and treat it as an inpainting task. This problem has been well studied using deep neural networks~\cite{yi2020contextual,yu2018generative} and patch matching based methods~\cite{bertalmio2000image,barnes2009patchmatch,huang2014image}. We use PatchMatch~\cite{barnes2009patchmatch} for its robustness. As demonstrated in the last two columns of Fig.~\ref{fig:result_table}, compared to directly regressing point-wise normal and texture, our inpainting-based results obtain sharper details and fewer artifacts.

\beforesection
\vspace{-2pt}
\section{Training Losses} \label{sec:learning}
\aftersection

The training process involves learning deep networks for two goals: joint-space occupancy estimation with $\mathcal{L}_{o}$, and normal/texture estimation with $\mathcal{L}_{n}$ and $\mathcal{L}_{t}$.
Specifically, $\mathcal{L}_{o}$ is the occupancy regression loss of our joint-space deep implicit functions, and $\mathcal{L}_{n},\mathcal{L}_{t}$ are image translation losses of the normal, texture estimation networks. 

\textbf{Joint-space Occupancy Estimation}.
The deep implicit function training is based on query point sampling and supervised occupancy regression with Tanh output layers.
We randomly sample mesh points $p_a$, $p_b$ in two spaces and then add diagonal Gaussian perturbation with a standard deviation of \SI{5}{cm} to increase the sample coverage of close-to-surface regions in space.
In each training iteration we sample 20480 pairs of query points $(p_a,p_b)$, with predicted occupancy $(o_a, o_b)$.
The joint-space occupancy regression loss contains three terms:
\begin{equation}
    \mathcal{L}_o(o_a, o_b) = \mathcal{L}_o^{occ}(o_a) + \mathcal{L}_o^{occ}(o_b) + \mathcal{L}_o^{con}(o_a, o_b),
\end{equation}
where $\mathcal{L}_o^{occ}(o_a), \mathcal{L}_o^{occ}(o_b)$ denote the Smooth $L1$-Loss between the estimated occupancy values and their ground truth in the canonical and the posed spaces, respectively. $\mathcal{L}_o^{con}(o_a,o_b)$ is a contrastive loss regularizing the occupancy consistency between the two spaces, that is,
\begin{equation}\small
\mathcal{L}_o^{con}(o_a, o_b)\!=\!
    \begin{cases}
    |o_a-o_b|,                              \!\!\!& \text{if } \hat{o}_a=\hat{o}_b,\\
    \lambda_1 \max(\lambda_2-|o_a-o_b|,0),  \!\!\!& \text{otherwise},
    \end{cases}
\end{equation}
where $\lambda_1$ and $\lambda_2$ are two parameters to adjust the penalty of inconsistent joint-space groundtruth pairs. Those pairs usually exist around the self-intersecting regions and need to be down-weighted due to the errors in canonical space supervision. Empirically, we set $\lambda_1 = 0.1$ and $\lambda_2 = 0.3$.

\begin{table*}[ptb]
\vspace{-2mm}
\centering
\setlength{\tabcolsep}{10pt}
\renewcommand\arraystretch{1.1}
\resizebox{\linewidth}{!}{
\begin{tabular}{@{}l||c|c|c||c|c|c||c|c|c}
\hline\thickhline
\multicolumn{1}{c||}{\multirow{2}{*}{\textbf{Components}}}  & \multicolumn{3}{c||}{\textbf{Posed Space}} & \multicolumn{3}{c||}{\textbf{Canonical Space}} & \multicolumn{3}{c}{\textbf{Mean}} \\
\cline{2-10} & Normal $\downarrow$ & P2S $\downarrow$ & Chamfer $\downarrow$ & Normal $\downarrow$ & P2S $\downarrow$ & Chamfer $\downarrow$ & Normal $\downarrow$ & P2S $\downarrow$ & Chamfer $\downarrow$ \\ 
\thickhline
Posed Sup. Only & 0.037 & 0.674 & 0.787 & 0.087 & 1.898 & 1.597 & 0.062     & 1.286    & 1.192   \\ 
Canonical Sup. Only & 0.039 & 0.716 & 0.838 & 0.046 & 0.606 & 0.997 & 0.043     & 0.661    & 0.917   \\ 
Joint & 0.037 & 0.662 & 0.789 & 0.045 & 0.620 & 0.988 & 0.041     & 0.641    & 0.825   \\ 
Joint + GeoEnc  & 0.033 & \textbf{0.495} & \textbf{0.614} & 0.040 & \textbf{0.471} & \textbf{0.819} & 0.036 & \textbf{0.483} & \textbf{0.717}   \\
Joint + GeoEnc + Refine & \textbf{0.031} & \textbf{0.495} & \textbf{0.614} & \textbf{0.039} & \textbf{0.471} & \textbf{0.819} & \textbf{0.035} & \textbf{0.483} & \textbf{0.717}   \\
\hline\thickhline
\end{tabular}}
\beforetab
\caption{\textit{Ablation studies on the effectiveness of ARCH++ proposed components in both spaces: posed vs. canonical}. Best scores are in \textbf{bold}.
Rows are target reconstruction spaces, columns are evaluation spaces. The first row means using the posed space as the target space (\textit{e.g.}, PIFu, PIFuHD, Geo-PIFu, PaMIR), whose reconstruction can be warped into the canonical space via a registered parametric body to compute evaluation metrics in both spaces. The second row means direct supervision and reconstruction in the canonical space, followed by warping into the posed space (\textit{e.g.}, ARCH). The rest rows are based on our joint-space co-supervision and reconstruction scheme.
}
\aftertab
\vspace{-1mm}
\label{tab:ablation_main}
\end{table*}

\textbf{Mesh Refinement}. We consider the image-space normal and texture estimation as an image-to-image translation task. Given an input image $I$, our task is to learn the front normal map $I_n^f$, the back normal map $I_n^b$ and the back side texture map $I_t^b \in \mathbb{R}^{512\times512\times3}$. Note we assume the input image can be directly used as the front texture map. Inspired by the demonstrated superior results of Pix2Pix~\cite{isola2017image,wang2018high}, we define the training losses as:
\begin{equation}\small
\begin{aligned}
    \mathcal{L}_{n}(I_n^f, I_n^b) &\!=\! \mathcal{L}_n^{rec}(I_n^f) \!+\! \mathcal{L}_n^{rec}(I_n^b) \!+\! \mathcal{L}_n^{vgg}(I_n^f) + \mathcal{L}_n^{vgg}(I_n^b),\\ 
    \mathcal{L}_{t}(I_t^b) &\!=\! \mathcal{L}_t^{rec}(I_t^b) + \mathcal{L}_t^{vgg}(I_t^b) + \mathcal{L}_t^{adv}(I_t^b),
\end{aligned}
\end{equation}
where $\mathcal{L}^{rec}(\cdot)$ denotes the $L1$ distance reconstruction loss, $\mathcal{L}^{adv}(\cdot)$ means the generative adversarial loss and $\mathcal{L}^{vgg}(\cdot)$ is the VGG-perceptual loss proposed by \cite{johnson2016perceptual}. In the experiments, we found that the generative adversarial loss $\mathcal{L}^{adv}(\cdot)$ counteracts to performance in the normal map estimation task and thus we only enforce this loss term upon the back side texture map.
One explanation is that the normal map space is more constrained and has fewer variations than the texture map, and therefore adversarial training does not fully show its effectiveness in this case.

\beforesection
\section{Experiments} \label{sec:experiments}
\aftersection

In this section, we present the experimental settings, result comparisons and ablation studies of ARCH++.

\beforesubsection
\subsection{Implementation Details}
\aftersection

We implement our framework using PyTorch and conduct the training with one NVIDIA Tesla V100 GPU. The proposed deep neural networks are trained with RMSprop optimizer with a learning rate starting from 1e-4. We use an exponential learning rate scheduler to update it every 3 epochs by multiplying with the factor \num{0.1} and terminate the training after 12 epochs.

\beforesubsection
\subsection{Datasets}
\aftersection

\begin{table}[ptb]
\centering
\setlength{\tabcolsep}{2pt}
\renewcommand\arraystretch{1.1}
\resizebox{\linewidth}{!}{
\begin{tabular}{@{}l||c|c|c||c|c|c@{}}
\hline\thickhline
\multicolumn{1}{c||}{\multirow{2}{*}{\textbf{Methods}}}  & \multicolumn{3}{c||}{\textbf{RenderPeople}} & \multicolumn{3}{c}{\textbf{BUFF}} \\
\cline{2-7}                             & Normal $\downarrow$ & P2S $\downarrow$ & Chamfer $\downarrow$ & Normal $\downarrow$ & P2S $\downarrow$ & Chamfer $\downarrow$\\ 
\thickhline
BodyNet~\cite{varol18_bodynet}          & 0.26      & 5.72    & 5.64     & 0.31     & 4.94    & 4.52   \\ 
VRN~\cite{VolumeRegECCVW2018}        	& 0.12      & 1.42    & 1.60     & 0.13     & 2.33    & 2.48   \\ 
SiCloPe~\cite{SiCloPeCVPR19}    	    & 0.22      & 3.81    & 4.02     & 0.22     & 4.06    & 3.99   \\ 
IM-GAN~\cite{chen2019implicit_decoder}  & 0.26      & 2.87    & 3.14     & 0.34     & 5.11    & 5.32   \\ 
PIFu~\cite{PIFuICCV19}      		    & 0.11      & 1.45    & 1.47     & 0.13     & 1.68    & 1.76   \\ 
PIFuHD~\cite{saito2020pifuhd}           & 0.11      & 1.37    & 1.43     & 0.13     & 1.63    & 1.75   \\ 
ARCH~\cite{huang2020arch}               & 0.04      & 0.74    & 0.85     & 0.04     & 0.82    & 0.87   \\ 
ARCH++ [Ours]                                    & \textbf{0.03}      & \textbf{0.50}    & \textbf{0.61}     & \textbf{0.03}     & \textbf{0.61}    & \textbf{0.64}   \\
\hline\thickhline
\end{tabular}}
\beforetab
\caption{\textit{Quantitative results and comparisons of normal, P2S and Chamfer errors between posed reconstruction and ground truth on RenderPeople and BUFF datasets}. Best scores are in \textbf{bold}.}
\aftertab
\label{tab:metrics}
\end{table}

We adopt the dataset setting from~\cite{huang2020arch,saito2020pifuhd}. Our training dataset consists of 375 3D scans from RenderPeople dataset~\cite{renderppl} and 205 3D scans from AXYZ dataset~\cite{axyz}.
These watertight human meshes have various clothes styles as well as body shapes and poses. Our testing set includes 37 scans from RenderPeople dataset~\cite{renderppl}, 192 scans from AXYZ dataset, 26 scans from BUFF dataset~\cite{zhang-CVPR17}, and 2D images from Internet public domains, representing clothed people with a large variety of complex clothes.
The subjects in the training dataset are mostly in standing pose, while the subjects in the test dataset
contain various poses including sitting, twisted and standing, as well as self-glued and separated limbs.
We use Blender and 38 environment maps to render each scan under different natural lighting conditions.
For each 3D scan, we generate 360 images by rotating a camera around the mesh with a step size of 1 degree.
These RenderPeople images are used to train both the occupancy estimation and the image translation networks. 

We generate ground truth clothed human meshes in the canonical pose using the method introduced in~\cite{huang2020arch}. Note that the warping process between the posed and the canonical spaces inevitably contain model noises (\eg, self-contact region artifacts, skinning weights nearest neighbor discontinuities), which motivates our joint-space co-supervision and reconstruction scheme.

\begin{figure}[ptb]
\centering
\includegraphics[width=\columnwidth,height=3cm]{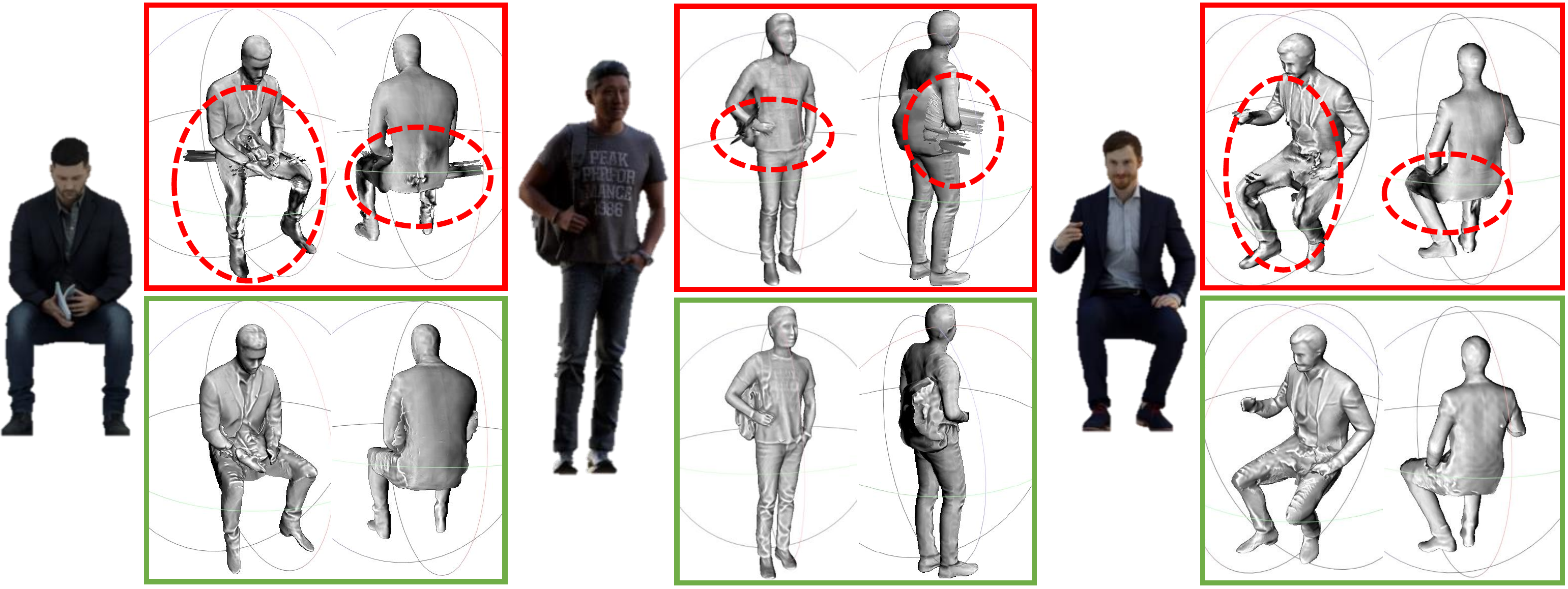}
\caption{We intentionally hide the method names for you to have a fair comparison on your own (\textit{please zoom in}). The answers are\protect\footnotemark.}
\label{fig:cmp_more_with_arch}
\vspace{-10pt}
\end{figure}
\footnotetext{Our results (green boxes) have fewer reconstruction artifacts (\eg, incorrect normal directions, mesh distortion) than ARCH (red boxes).}

\begin{figure*}[ht]
\vspace{-2mm}
\centering
\includegraphics[width=\textwidth]{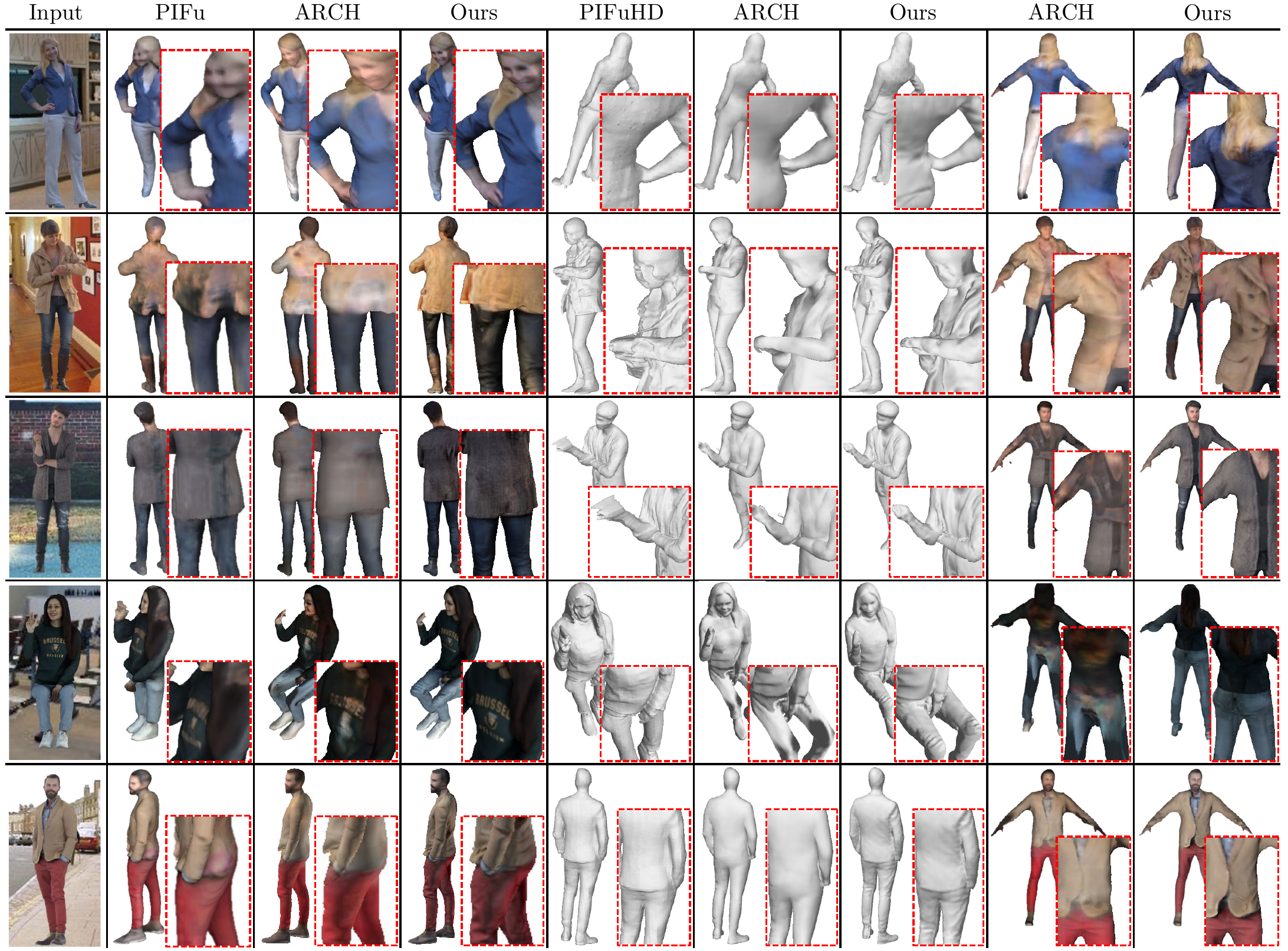}
\beforefigcaption
\caption{\textit{Qualitative comparisons against the state-of-the-art methods~\cite{PIFuICCV19,saito2020pifuhd,huang2020arch}.}
The first column is input. Column 2-4, 5-7 are color and shape reconstruction results, respectively, in the posed space. The last two columns are canonical space avatar reconstructions.
Our method handles arbitrary poses with self-contact and occlusions robustly, and reconstructs a higher level of details than existing methods.
}
\afterfigcaption
\label{fig:result_table}
\end{figure*}

\beforesubsection
\subsection{Results and Comparisons} \label{sec:result}
\aftersection

We use the same metrics as~\cite{PIFuICCV19,saito2020pifuhd,huang2020arch} for quantitative evaluation of the reconstructed meshes. We report the average point-to-surface Euclidean distance (P2S) and the Chamfer distance in centimeters, as well as the $L2$ normal re-projection errors. The two state of the art methods for our main comparisons are PIFuHD~\cite{saito2020pifuhd} and ARCH~\cite{huang2020arch}, both are built upon PIFu~\cite{PIFuICCV19} with improvements in different aspects. PIFuHD ingests high-resolution images in a sliding window manner to achieve rich surface reconstruction details. ARCH leverages nearest neighbor-based linear blend skinning weights and hand-crafted RBF features to reconstruct animatable avatars in a canonical space. In addition to these two most related methods, we also include multiple prior methods~\cite{varol18_bodynet,VolumeRegECCVW2018,SiCloPeCVPR19,chen2019implicit_decoder,PIFuICCV19} and report the benchmark results on the RenderPeople and the BUFF datasets in Tab.~\ref{tab:metrics}. ARCH++~[Ours] results outperform the second best method ARCH by large gaps.

\begin{figure}[ptb]
\vspace{-1mm}
\centering
\includegraphics[width=\linewidth]{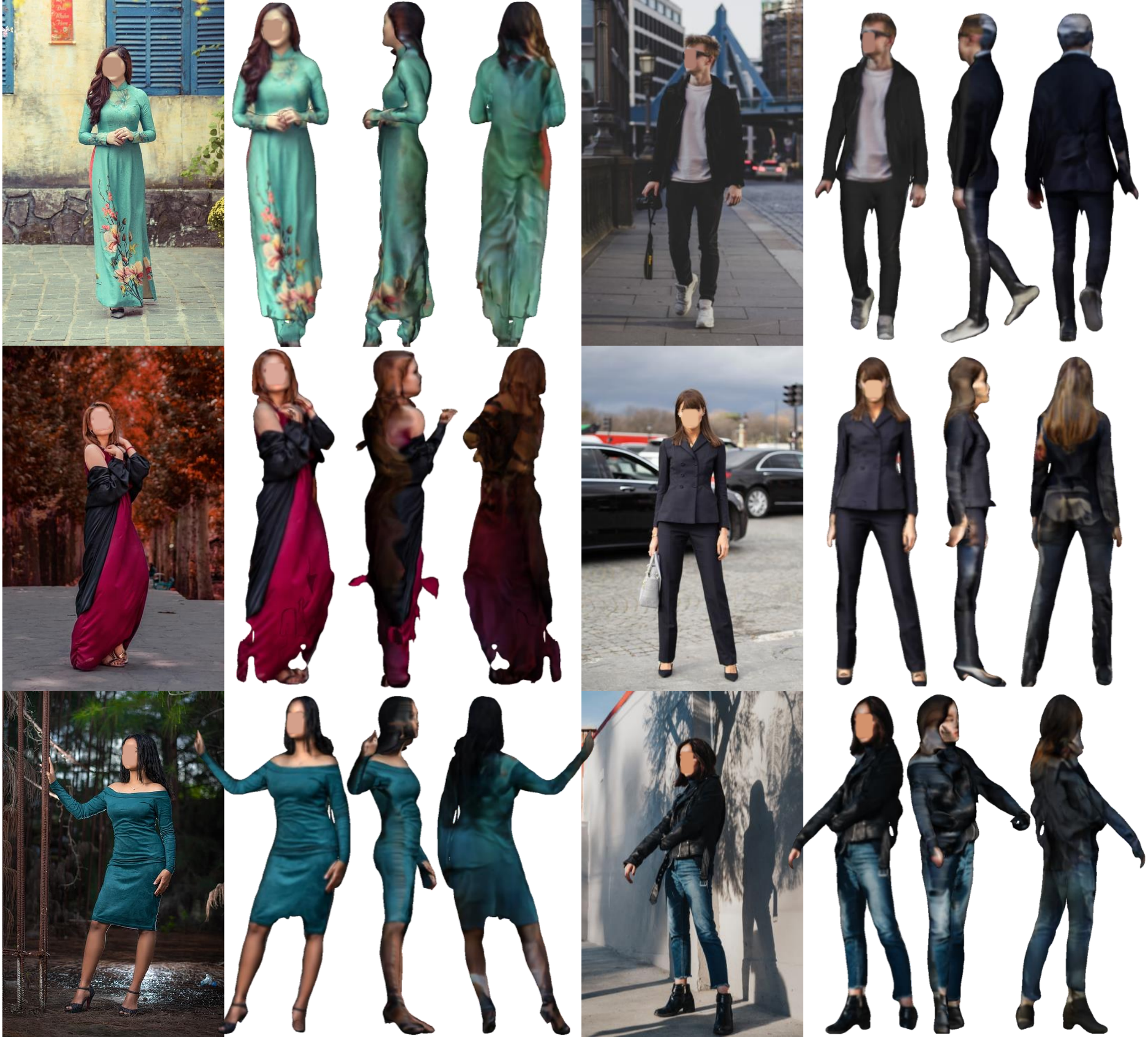}
\beforefigcaption
\caption{\textit{An application of digital human capture from photos}.}
\afterfigcaption
\label{fig:in_the_wild}
\end{figure}

The visual comparisons in Fig.~\ref{fig:result_table} and Fig.~\ref{fig:cmp_more_with_arch} further explain the advantages of our improvement. PIFuHD suffers from shape distortions due to lacking shape and pose priors provided by the end-to-end geometry encoder. Note that PIFuHD is incapable of reconstruct canonical space avatars and lacks texture estimation. ARCH reconstructions tend to be over smooth and blurry. Its recovered mesh normal and texture also have several block artifacts. Additionally, both methods fail to hallucinate plausible back-side surface details like clothes wrinkles, hairs, \textit{etc}. In comparison, our approach achieves photorealistic and animatable reconstructions in joint spaces and across different viewpoints. We further show our results on Internet images in Fig.~\ref{fig:in_the_wild}.

\subsection{Ablation Studies}\label{sec:ablation}

\begin{table}
\centering
\setlength{\tabcolsep}{12pt}
\renewcommand\arraystretch{1.1}
\resizebox{\linewidth}{!}{
\begin{tabular}{@{}l||c|c|c}
\hline\thickhline
\multicolumn{1}{c||}{\multirow{1}{*}{\textbf{Variants}}} & \multicolumn{1}{c|}{Normal $\downarrow$}  & \multicolumn{1}{c|}{P2S $\downarrow$}     & \multicolumn{1}{c}{Chamfer $\downarrow$}   \\ 
\thickhline
Depth~\cite{PIFuICCV19}                 & 0.047   & 0.78     & 0.93 \\
RBF~\cite{huang2020arch}                & 0.042   & 0.74     & 0.85 \\
End-to-end Voxel~\cite{he2020geopifu,Zerong2020PaMIR}   & 0.034   & 0.52     & 0.63 \\
End-to-end Point  & \textbf{0.033}     & \textbf{0.50}     & \textbf{0.61} \\ 
\hline\thickhline
\end{tabular}}
\beforetab
\caption{\textit{Ablation studies on different types of geometry encoders}.
}
\aftertab
\label{tab:ablation_geo}
\end{table}

\begin{figure}[ptb]
\vspace{-1mm}
\centering
\includegraphics[width=\linewidth]{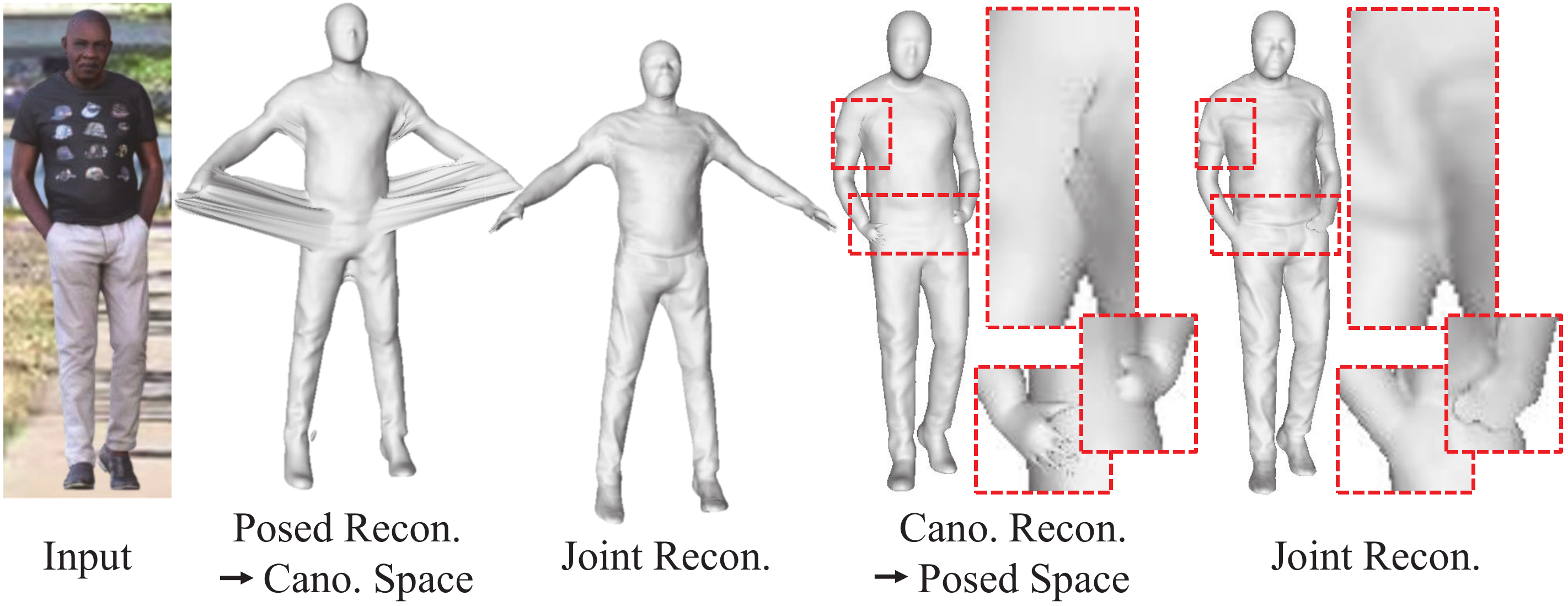}
\beforefigcaption
\vspace{1mm}
\caption{\textit{Ablation studies on the reconstruction space}. Single-space reconstruction shows artifacts of either mesh surface over-stretching or intersecting surfaces when warping from one space to another. Our joint-space reconstruction obtains balanced performance of both high reconstruction completeness under the canonical space and high input image fidelity under the posed space.
}
\afterfigcaption
\vspace{1mm}
\label{fig:ablation_space}
\end{figure}

\textbf{Joint Space Reconstruction}. To further understand the impact of the proposed methods, we present ablation studies in Tab.~\ref{tab:ablation_main}. The first three rows demonstrate the effectiveness of joint-space co-supervision, achieving balanced performances on both the posed and the canonical space mesh reconstructions. Choosing the posed space as the reconstruction target space (\eg, PIFu, PIFuHD, Geo-PIFu, PaMIR) can cause missing surfaces and topology distortions in the posed-to-canonical space warped meshes (see Fig.~\ref{fig:ablation_space}). Meanwhile, choosing the canonical space as the target space (\eg, ARCH) can cause self-intersecting meshes with broken manifold as well as body part un-natural deformations in the canonical-to-posed space warped meshes. In contrast, our co-supervision and joint-space inference methods achieve both reconstruction fidelity in the posed space and body mesh completeness in the canonical space.

\begin{figure}[ptb]
\centering
\includegraphics[width=\linewidth]{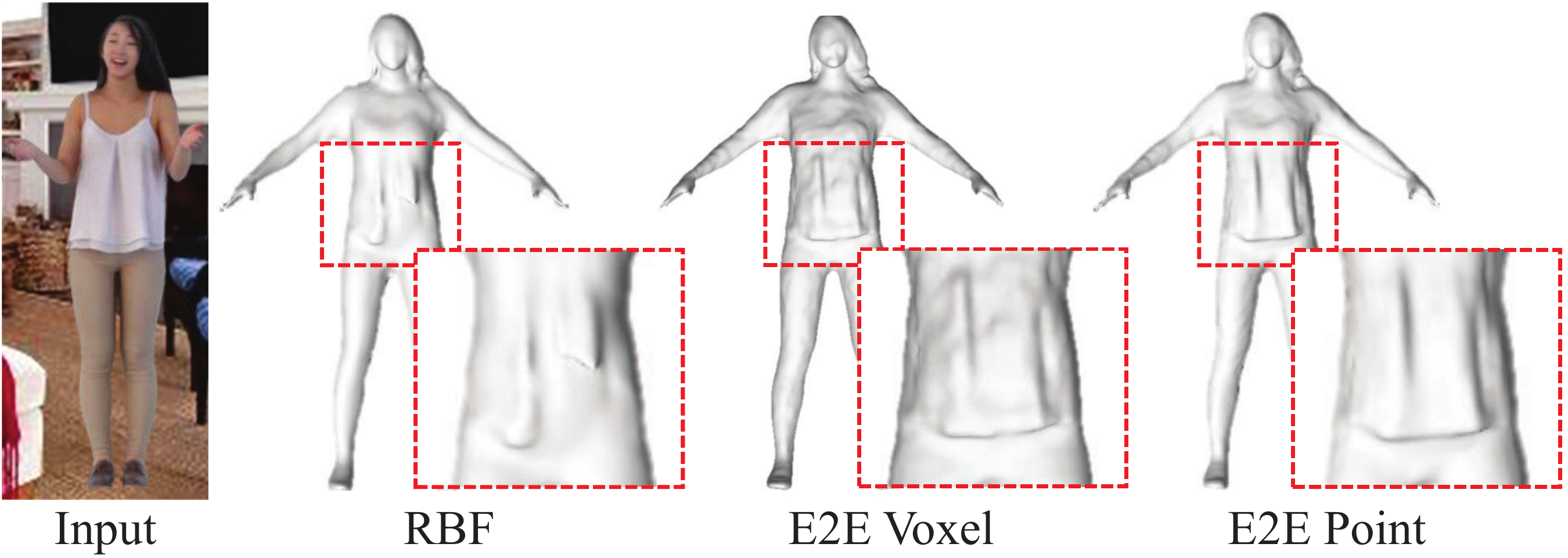}
\beforefigcaption
\caption{\textit{Ablation studies on geometry encoding}. Learned spatial features capture both pose and shape priors of the underlying parametric models and thus enable mesh reconstruction with more surface details than the handcrafted RBF features. Meanwhile, results of the voxel-based features are noisier than the point-based ones due to mesh quantization (\ie, voxelization) errors.}
\afterfigcaption
\vspace{1mm}
\label{fig:ablation_geo}
\end{figure}

\begin{figure}[ptb]
\centering
\includegraphics[width=\linewidth]{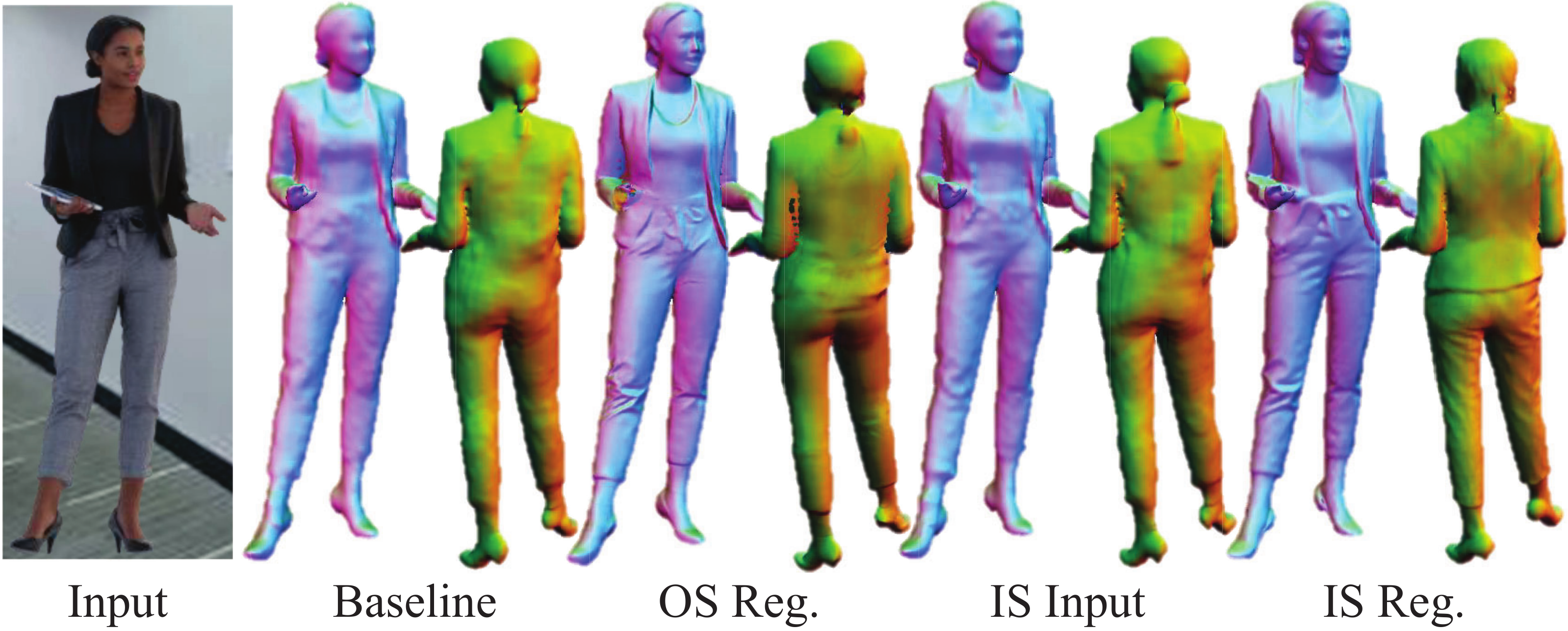}
\beforefigcaption
\caption{\textit{Ablation studies on normal refinement}: Object-space Regression (OS Reg.), Image-space Input (IS Input) and Image-space Regression (IS Reg.). Our method IS Reg. leads to rich reconstruction details (\eg, clothes wrinkles) in all views.
}
\afterfigcaption
\vspace{1mm}
\label{fig:ablation_normal}
\end{figure}

\begin{table}[ptb]
\centering
\setlength{\tabcolsep}{6pt}
\renewcommand\arraystretch{1.0}
\resizebox{\linewidth}{!}{
\begin{tabular}{@{}l||c|c|c}
\hline\thickhline
\multicolumn{1}{c||}{\multirow{1}{*}{\textbf{Variants}}}  & \multicolumn{1}{c|}{Posed $\downarrow$} & \multicolumn{1}{c|}{Canonical $\downarrow$} & \multicolumn{1}{c}{Mean $\downarrow$} \\
\thickhline
Baseline                                        & 0.033 & 0.040 & 0.037   \\
Object-space Regression~\cite{huang2020arch}    & 0.032 & 0.041 & 0.037   \\
Image-space Input~\cite{saito2020pifuhd}        & 0.032 & \textbf{0.038} & \textbf{0.035}   \\
Image-space Regression                          & \textbf{0.031} & 0.039 & \textbf{0.035}   \\
\hline\thickhline
\end{tabular}}
\beforetab
\caption{\textit{Ablation studies on different ways of normal refinement}. 
}
\aftertab
\label{tab:ablation_normal}
\end{table}

\textbf{Geometry Encoding}. As shown in Tab.~\ref{tab:ablation_main}, we observe further error reduction leveraging the end-to-end learned point-wise spatial encodings. The prior method ARCH uses handcrafted RBF features that only model the pose prior of parametric body mesh skeletons, ignoring the mesh shape. In comparison, our point-based features are informed of both pose and shape priors of the underlying parametric body model \wrt a clothed human mesh, and thus improve the surface reconstruction quality. We further implement the learned volumetric spatial feature encodings used in Geo-PIFu and PaMIR as an alternative encoder and inject into our framework for direct comparisons. The results are shown in Tab.~\ref{tab:ablation_geo} and Fig.~\ref{fig:ablation_geo}. While both types of end-to-end spatial features outperform the hand crafted RBF features, our point-based feature extraction method does not suffer from computation overhead and mesh quantization errors of the voxel-based approach.

\textbf{Normal Refinement}. While single-image based direct inference of human meshes with rich surface details at both the front and the back side remains an open question, some empirical observations and prior works indicate that normal estimation is a relatively easier task and can help refine the reconstructions. In Tab.~\ref{tab:ablation_normal} and Fig.~\ref{fig:ablation_normal} we experiment on three principle ways of leveraging the estimated normals for mesh reconstructions with refined surface details. Among these normal refinement methods, our front/back-side image space normal regression and moulding-based surface refinement approach outperforms other variants. Object-space normal regression is adopted in ARCH and is based on learning deep implicit functions of spatial normal fields. It fails to generate rich back side details and sometimes causes block artifacts as shown in the fourth row of Fig.~\ref{fig:result_table}. Image-space input is used in PIFuHD. It concatenates the color image input with estimated image-space normal maps and feeds them into Stack Hourglass for feature extraction. While this method achieves the same level of quantitative performance as our mesh refinement approach, its visual results are not as sharp as ours at both the front and the back sides. A degenerated case of our mesh refinement method is studied before in DeepHuman where they only estimate front-view normal maps and therefore lack reconstruction details at the back side.

\beforesection
\section{Conclusion} \label{sec:conclusion}
\aftersection

In this paper, we revisit the major components in existing deep implicit function based 3D avatar reconstruction. Our method ARCH++ produces results which have high-level fidelity and are animation-ready for many AR/VR applications. We conduct a series of comparisons with and analysis on the state of the art to validate our findings. For future works, we plan to incorporate environment information (e.g., lighting, affordance) to further understand the body pose and appearance, and address current limitations.

\noindent {\small{\textbf{Acknowledgements.}} We would like to thank Minh Vo and Nikolaos Sarafianos for the discussions and synthetic data creation.}

{\small
\bibliographystyle{ieee_fullname}
\bibliography{3d_human}
}

\clearpage

\appendix

\section*{Supplementary}

\section{Implementation Details}

In this section, we provide the implementation details of our proposed method.

\subsection{Input Preprocessing}

During both training and test time, the input images to the network are normalized with regard to the human body scale. In particular, we re-scale the image based on the 3D skeleton estimation of the subject.
The image is resized then centered, such that the pelvis of the person is aligned with the center of the image. Each pixel represents \num{1}cm length using an orthographic scene projection. In this way we ensure proper scaling of the body parts, which allows us to capture the variations of different heights of people.

\subsection{Network Architectures}

\textbf{Semantic-Aware Geometry Encoder} is based on PointNet++~\cite{qi2017pointnet++,qi2017pointnet}, which consists of 3 Set Abstraction (SA) layers. The configurations of each layer are SA(2048, 0.1, 16, 3, [16,16,32]), SA(512, 0.2, 32, 32, [32,32,64]), SA(128, 0.4, 64, 64, [64,64,128]). The explanation of each argument is (furthest point sampling size, point neighborhood radius, point neighborhood size limit, input feature channel, MLP output channels list). Namely, the multi-scale point set sizes of Eq. (1) in the main paper are: $N_1=2048, N_2=512, N_3=128$. When extracting spatially-aligned features for any given query point, we leverage the point Feature Propagation (FP) layers defined at the aforementioned 3 different point set scales: FP(32, [32,32]), FP(64, [32,32]), FP(128, [32,32]). The explanation of each argument is (input feature channel, MLP output channels list). Therefore, the dimensions of our spatially-aligned geometry features $f_g$ in Eq. (3) of the main paper are $96 = 32*3$. Please refer to~\cite{qi2017pointnet++} for further details.

\textbf{Pixel-Aligned Appearance Encoder} adopts the architecture from Stack Hourglass Network~\cite{newell2016stacked}. The layer configuration is the same as PIFu(HD) and ARCH~\cite{PIFuICCV19,saito2020pifuhd,huang2020arch}, which is composed of a 4-stack model and each stack uses 2 residual blocks. The output latent image feature length is 256. Therefore, the dimensions of our pixel-aligned appearance features $f_a$ in Eq. (3) of the main paper are $256$. Please refer to~\cite{newell2016stacked,PIFuICCV19,saito2020pifuhd,huang2020arch} for further details.

\textbf{Joint-Space Occupancy Estimator} is a two-branch multilayer perceptron (MLP). Each branch takes the spatially-aligned geometry features $f_g$ and pixel-aligned appearance features $f_a$ described above, and estimates one-dimension occupancy $o_a$ or $o_b$ using Tanh activation. $o_a$ is the occupancy probability in the canonical space and $o_b$ is the one in the posed space. Similar to~\cite{PIFuICCV19}, we design the MLP with four fully-connected layers and the numbers of hidden neuron sizes are $(1024,512,256,128)$. Each layer of MLP has skip connections from the input features.

\textbf{Normal and Texture Image Translation} both use a network architecture designed by~\cite{johnson2016perceptual} using 9 residual blocks with 4 downsampling layers. The same network is also used in PIFuHD~\cite{saito2020pifuhd} for normal maps estimation. In our work we extend this architecture to back side texture inference by adding GAN losses.

\subsection{Hyper-Parameters}
When training the joint-space occupancy losses $\mathcal{L}_o$, we use 0.5, 0.5, 0.05 to weight the canonical/posed space occupancy estimation losses $\mathcal{L}_o^{occ}$ as well as the contrastive regularizer $\mathcal{L}_{o}^{con}$. When supervising the normal and texture image translation losses $\mathcal{L}_n, \mathcal{L}_t$, we set the weights of the $L1$ reconstruction losses $\mathcal{L}^{rec}$ and the perceptual losses $\mathcal{L}^{vgg}$ to 5.0 and 1.0, respectively. Particularly, the GAN losses (\textit{i.e.} generator, discriminator) used in back-side texture hallucination is weighted by 0.1.

\begin{figure}[ptb]
\centering
\includegraphics[width=\linewidth]{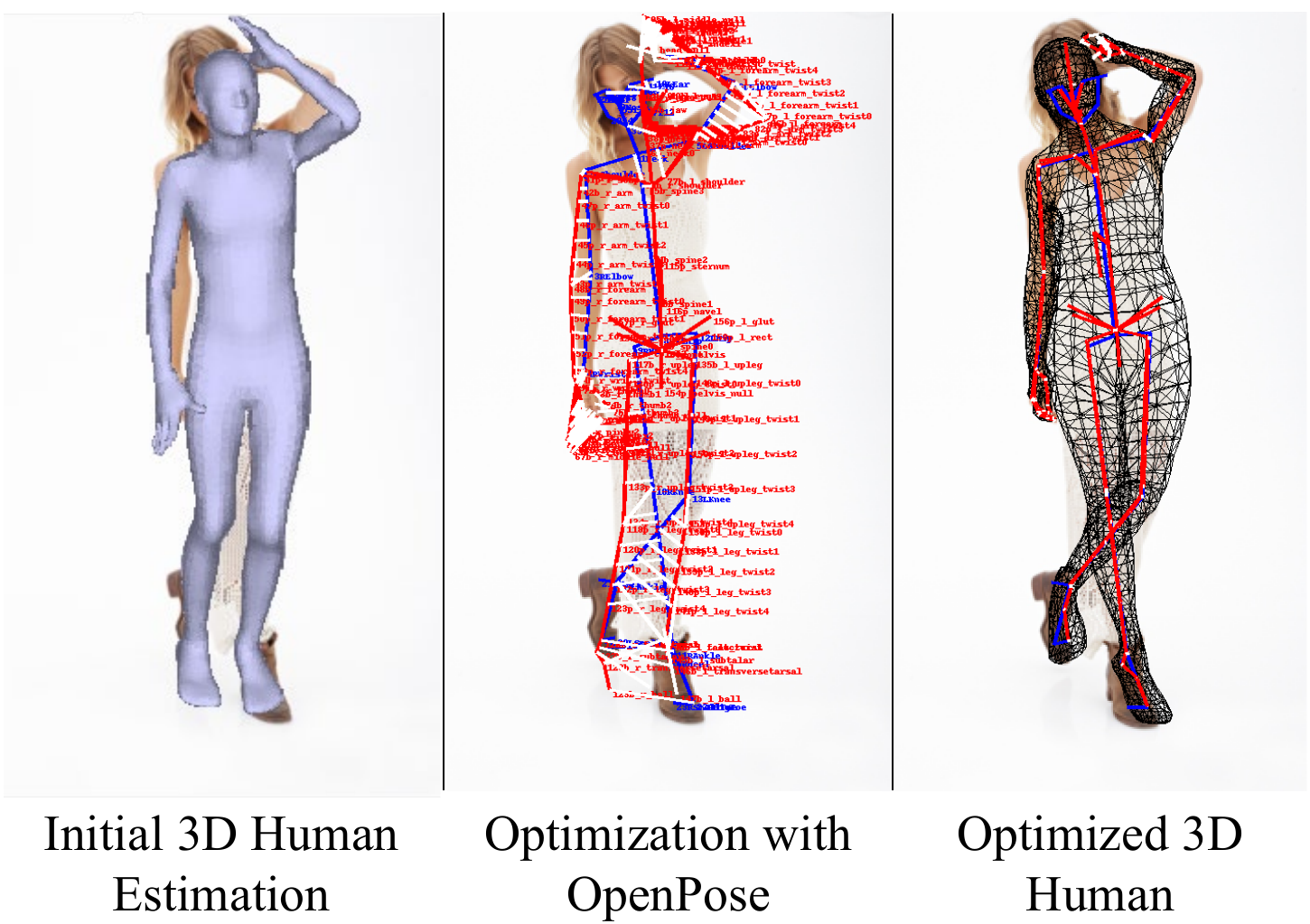}
\beforefigcaption
\vspace{+3pt}
\caption{\textit{Postprocessing of 3D human body estimations}. We observe the initially estimated 3D human body does not perfectly align with the 2D body landmarks (\eg, face, hands, feet) and we thus implement an additional optimization script to further minimize the distance between the re-projected 2D landmarks (red) and OpenPose detected 2D landmarks (blue), leading to a better 3D human body estimation.
}
\vspace{+4pt}
\afterfigcaption
\label{fig:denserac_post}
\end{figure}

\subsection{Computation Cost}

Empirically, when using a single Tesla V100 GPU for training and one batch of \num{4} images (each image with 20480 pairs of query points), the forward pass takes around 1.4s and the backward propagation takes around 0.6s.

\begin{figure*}[ptb]
\vspace{-30 pt}
\centering
\includegraphics[width=0.9\linewidth]{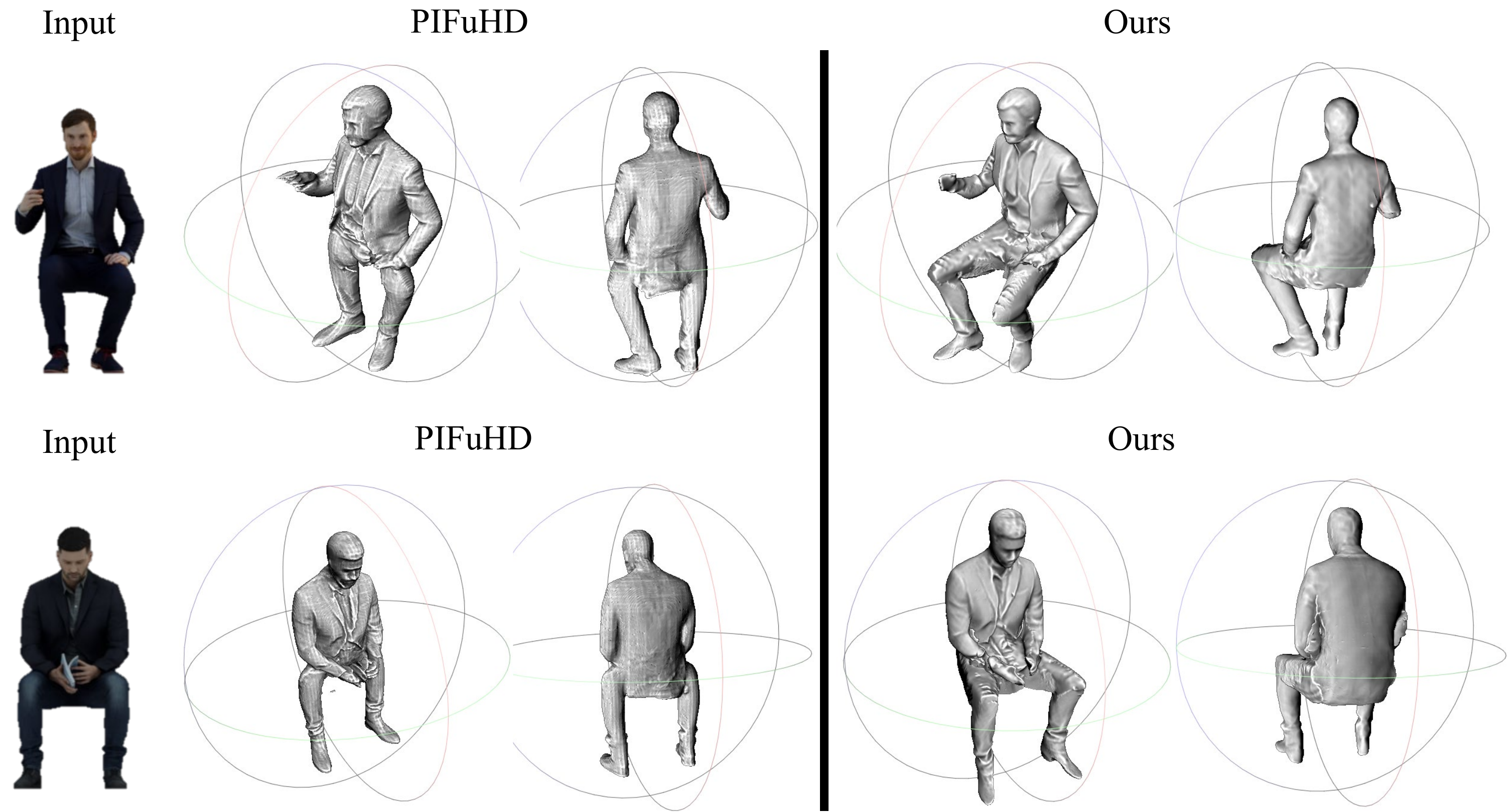}
\beforefigcaption
\vspace{+10pt}
\caption{\textit{3D human body prior is crucial for our task}, without which the reconstructions might look like squeezed relief sculpture due to wrong poses and shapes.
}
\afterfigcaption
\label{fig:body_priors_matter}
\end{figure*}

\section{Inference on Images in the Wild}

To perform inference on in-the-wild images, our staged pipeline involves person instance segmentation, parametric 3D human body estimation and the proposed 3D avatar reconstruction. The total pipeline takes around \num{5} seconds to reconstruct a fully colored animation-ready avatar from an unconstrained photo (RGB image) using one Tesla V100 GPU. In comparison, PaMIR \cite{Zerong2020PaMIR} takes over \num{40} seconds. Note our current implementation lets all modules run sequentially and the intermediate results (\eg, masks, 3D human body parameters) are mostly exchanged through CPU memory and file IO, which leaves room for optimization. For our proposed avatar reconstruction framework alone, we could run 1) Semantic-Aware Geometry Encoder, 2) Pixel-Aligned Appearance Encoder and 3) Normal and Texture Refinement Networks in parallel to greatly boost the efficiency. 

\subsection{Person Instance Segmentation}

Similar to most existing implicit surface function based methods, our occupancy estimation module also requires the person segmentation mask. Such mask is used to remove redundant and erroneous estimated occupancy in the background regions and serves as a visual hull prior similar to multiple view stereo. In this paper, we utilize one state of the art semantic instance segmentation~\cite{pointrend20}, which is able to generate per-person segmentation mask. Note we are able to handle multiple people in the same image with such a detection-and-segmentation method. We set a minimum detection score \num{0.5} and minimum bounding box size $100 \times 100$ to filter out people instances with too small resolution and guarantees the proper scale used for our proposed avatar reconstruction approach.

\subsection{Parametric 3D Human Body Estimation}


Underlying 3D human body serves as an important semantic cue for our approach. In this paper, we adopt a similar way as ARCH~\cite{huang2020arch} to estimate the parametric 3D human body from the input image. We first run DenseRaC~\cite{DenseRaCICCV19} to obtain the initial estimation of 3D human pose and shape parameters. Furthermore, we observe that when re-projecting such estimated 3D human body back to the input image, the body landmarks (\eg, joints, face, hands, feet) do not align with the input image well. We thus implement an additional optimization script using pytorch to compute the offsets between the re-projected body landmarks and detected body landmarks from OpenPose~\cite{openpose} and back-propagate to the estimated 3D human pose and shape parameters (see Fig.\ref{fig:denserac_post}). The optimization is run over 200 iterations and we obtain better-aligned 3D human body in this way.


\subsection{3D Avatar Reconstruction}

Given the intermediate results obtained from the modules above, we are able to run our proposed approach and obtain the jointly reconstructed avatars in both the original posed space and the canonical space.

\section{Extended Experiments}
In this section, we show some interesting conclusions we obtained along the way and extended experiments as well as comparisons (\eg, user studies, applications of avatar animations and video-based fusion, failure cases). Note we remove the background for all images for better visualization of the inputs.


\subsection{3D Human Body Prior is Crucial for Our Task}

In Fig.~\ref{fig:body_priors_matter} we show that reconstruction results of PIFuHD~\cite{saito2020pifuhd} fail to capture the underlying correct body shapes and poses. PIFuHD and our method are trained using the same set of RenderPeople clothed human meshes, which consist of mostly upstanding poses. While obtaining large-scale ground truth clothed avatars with various poses and shapes is still an open problem, we can leverage parametric body shape estimation networks (\textit{e.g.} DenseRaC~\cite{DenseRaCICCV19} and HMR~\cite{kanazawa2018hmr}) whose training data is easier to obtain. This motivates our design of learning both semantic-aware geometry features and pixel-aligned appearance features. The geometry features encode shape and pose priors of the underlying parametric body mesh, while the appearance features provide image evidence for fine-scale clothing wrinkles and surface details reconstruction.

\begin{figure}[ptb]
\centering
\includegraphics[width=\linewidth]{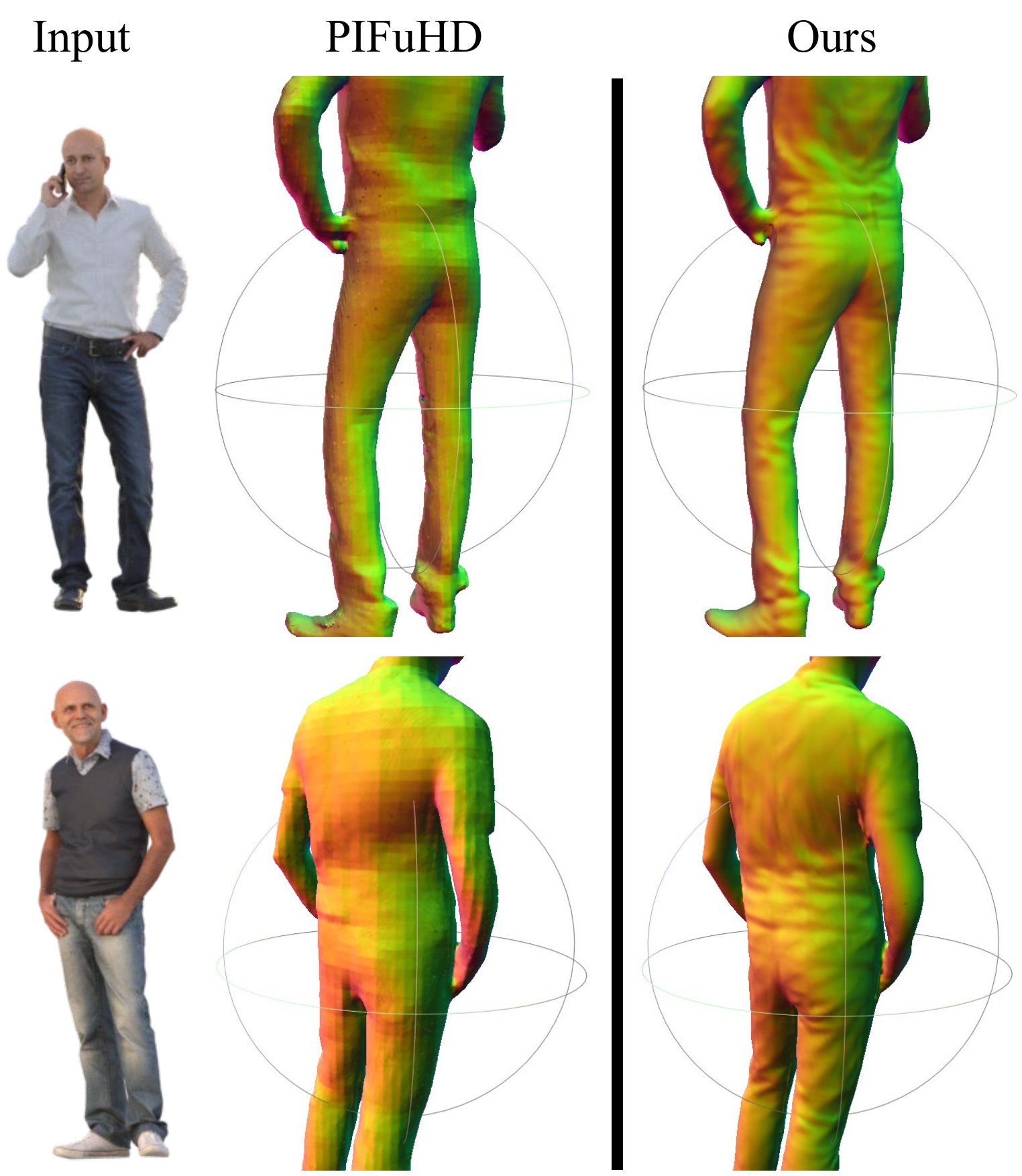}
\beforefigcaption
\vspace{+3pt}
\caption{\textit{Moulding-based Surface Refinement Obtains Better Consistency across Views.} Our reconstruction results contain more full-body surface details (\eg, belt around the waist and clothing wrinkles on the back/legs) than the competing methods.
}
\afterfigcaption
\label{fig:supp_zoom_in_back}
\vspace{+5 pt}
\end{figure}

\begin{figure}[ptb]
\centering
\includegraphics[width=\linewidth]{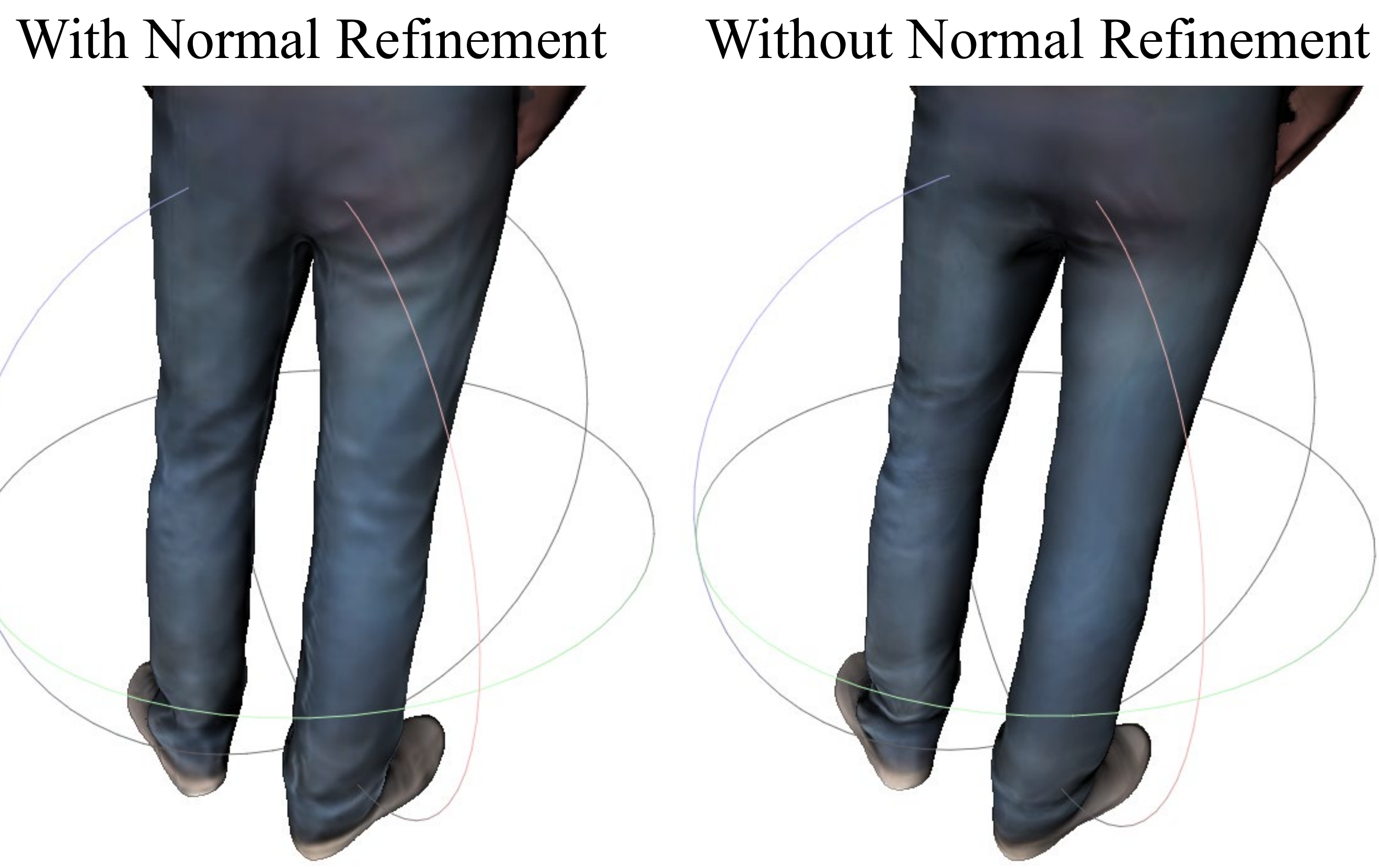}
\beforefigcaption
\vspace{+3pt}
\caption{\textit{Normal refinement can enhance photorealistic rendering at different viewpoints.} These images use the same mesh textures for rendering in order to demonstrate the effect of refined normals. The input image is the same as the bottom row in Fig.~\ref{fig:supp_zoom_in_back}. 
}
\afterfigcaption
\label{fig:normal_shading}
\end{figure}

\subsection{Moulding-based Surface Refinement Obtains Better Consistency across Views}

For previous methods like PIFu~\cite{PIFuICCV19}, ARCH~\cite{huang2020arch} and PIFuHD~\cite{saito2020pifuhd}, we often observe the reconstructed surface details only look plausible from the input camera view. Once we change the camera view to preview the reconstructed avatar from other view points, the rendered results contain fewer surface details and are less realistic. Such quality inconsistency limits the applicability of the prior works to AR/VR applications that require free viewpoint rendering.

Based on the aforementioned observations, we conclude that such phenomenon is caused by the suppression of occluded region hallucination. Although PIFuHD generates detailed back-side surface compared to PIFu and ARCH by leveraging inferred normal maps, its final rendering quality remains less sharp than our moulding-based refinement approach. Besides the results shown in the main paper, we provide more results with zoom-in in Fig.~\ref{fig:supp_zoom_in_back} to further demonstrate the improvement on reconstruction details.
Moreover, the normal refinement step enhances photorealistic rendering results by enabling fine-grained shading effects. In Fig.~\ref{fig:normal_shading}, we show the rendered images with and without the refined normals using the same mesh textures. With the normal refinement, the rendered images show more plausible clothing wrinkles at different views than the ones without normal refinement.

\begin{figure*}[ptb]
\vspace{-15 pt}
\centering
\includegraphics[width=\linewidth]{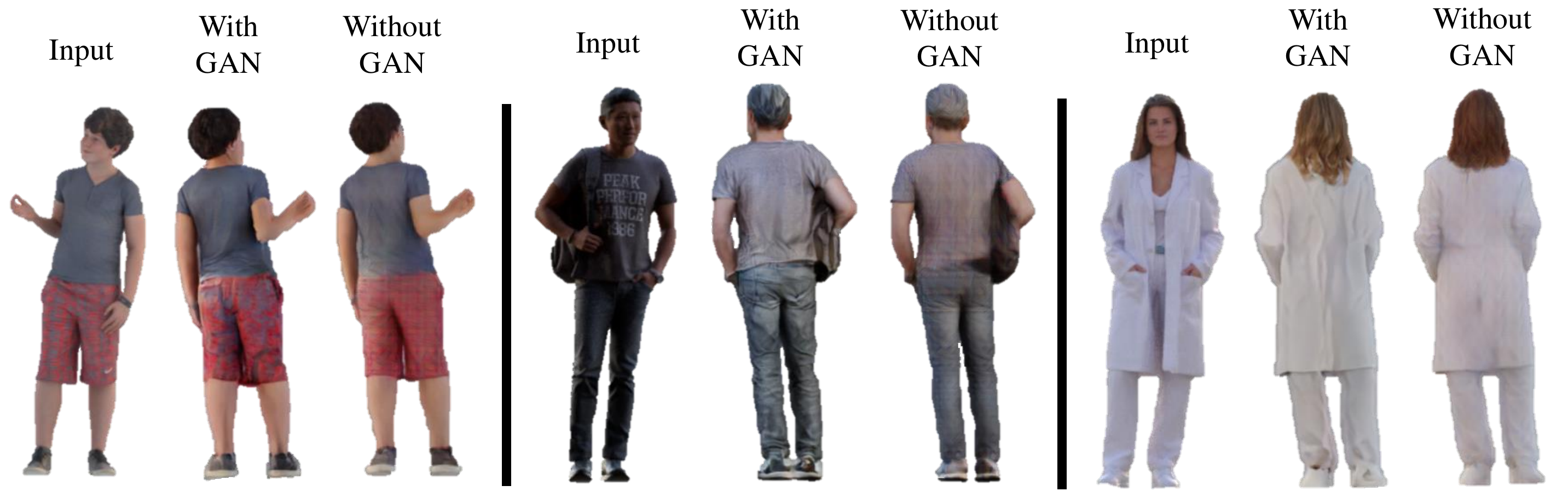}
\beforefigcaption
\caption{\textit{Texture Map Estimation With and Without GAN}. The hallucinated back-side texture maps with GAN contain more surface details (\eg, clothing patterns/wrinkles, hairs) and more realistic (directional) lighting effect than those without GAN.
}
\vspace{+10pt}
\afterfigcaption
\label{fig:supp_texture_gan}
\end{figure*}

\begin{figure*}[ptb]
\centering
\includegraphics[width=\linewidth]{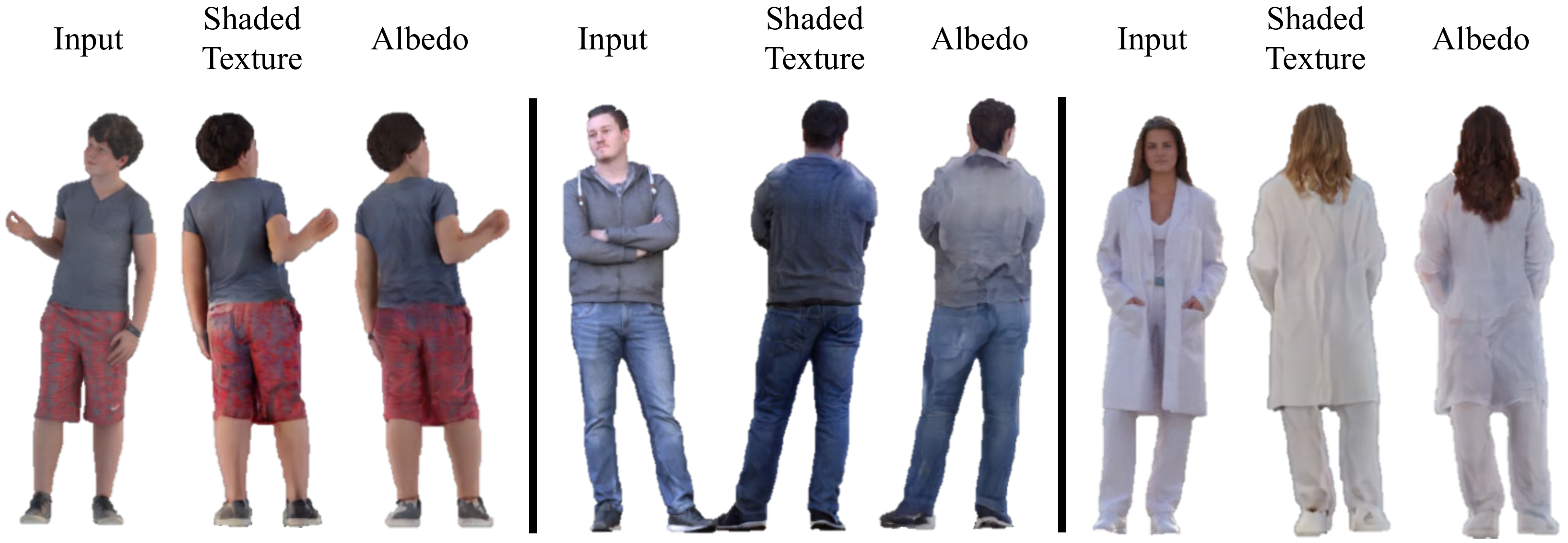}
\beforefigcaption
\caption{\textit{Texture Map Estimation With and Without Shading (Albedo)}. We study whether the shaded texture or the albedo space are easier to learn. Results show when trying to predict albedos, we lose lots of clothing details and textured patterns.
}
\afterfigcaption
\label{fig:supp_albedo_gan}
\end{figure*}

\begin{figure*}[ptb]
\centering
\includegraphics[width=\linewidth]{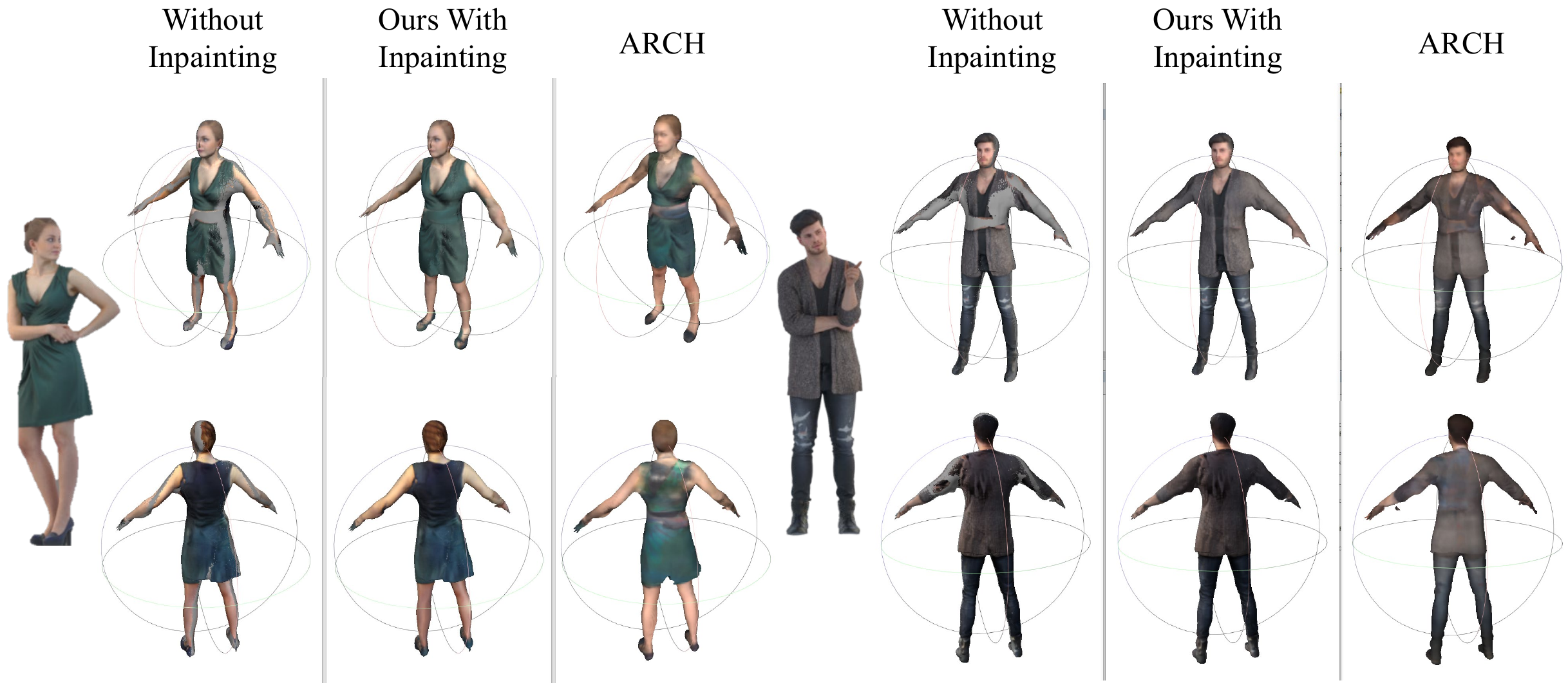}
\beforefigcaption
\caption{\textit{Inpainting Is a Needed Step for Avatar Reconstruction}. Canonical space reconstruction inevitably contains unseen surfaces from the input image. After applying a ray tracing and moulding, we identify those gray regions can be seen in neither front side nor back side as occluded region and fill in those missing details using an off-the-shelf image inpainting algorithm. Compared with ARCH which interpolates the per-point normal/texture using deep implicit functions, our moulding-inpainting approach obtains higher fidelity and completeness.
}
\afterfigcaption
\label{fig:inpaint}
\vspace{+15 pt}
\end{figure*}

\begin{figure*}[ptb]
\centering
\includegraphics[width=\linewidth]{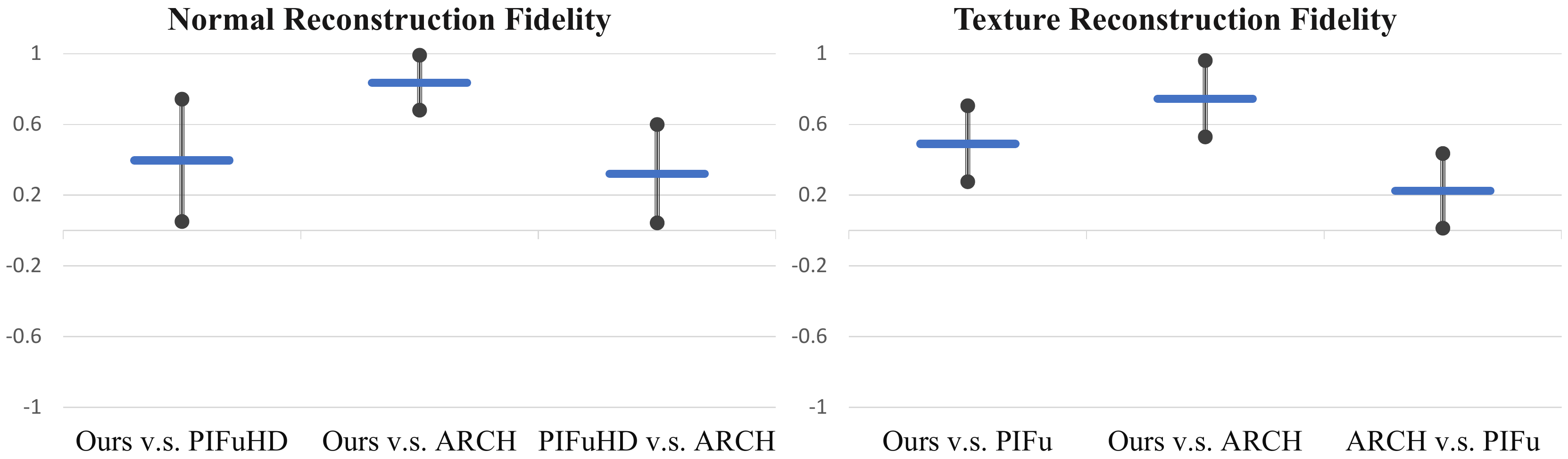}
\caption{Results of our user study. Values $>0$ show a preference of the first method over the second, \ie, Ours over PIFuHD, Ours over ARCH and PIFuHD over ARCH. Note PIFuHD doesn't reconstruct texture so we use PIFu instead for the texture reconstruction study. Error bars show standard error.}
\label{fig:user_study}
\vspace{-8 pt}
\end{figure*}


\begin{figure*}[ptb]
\vspace{+6 pt}
\centering
\includegraphics[width=\linewidth]{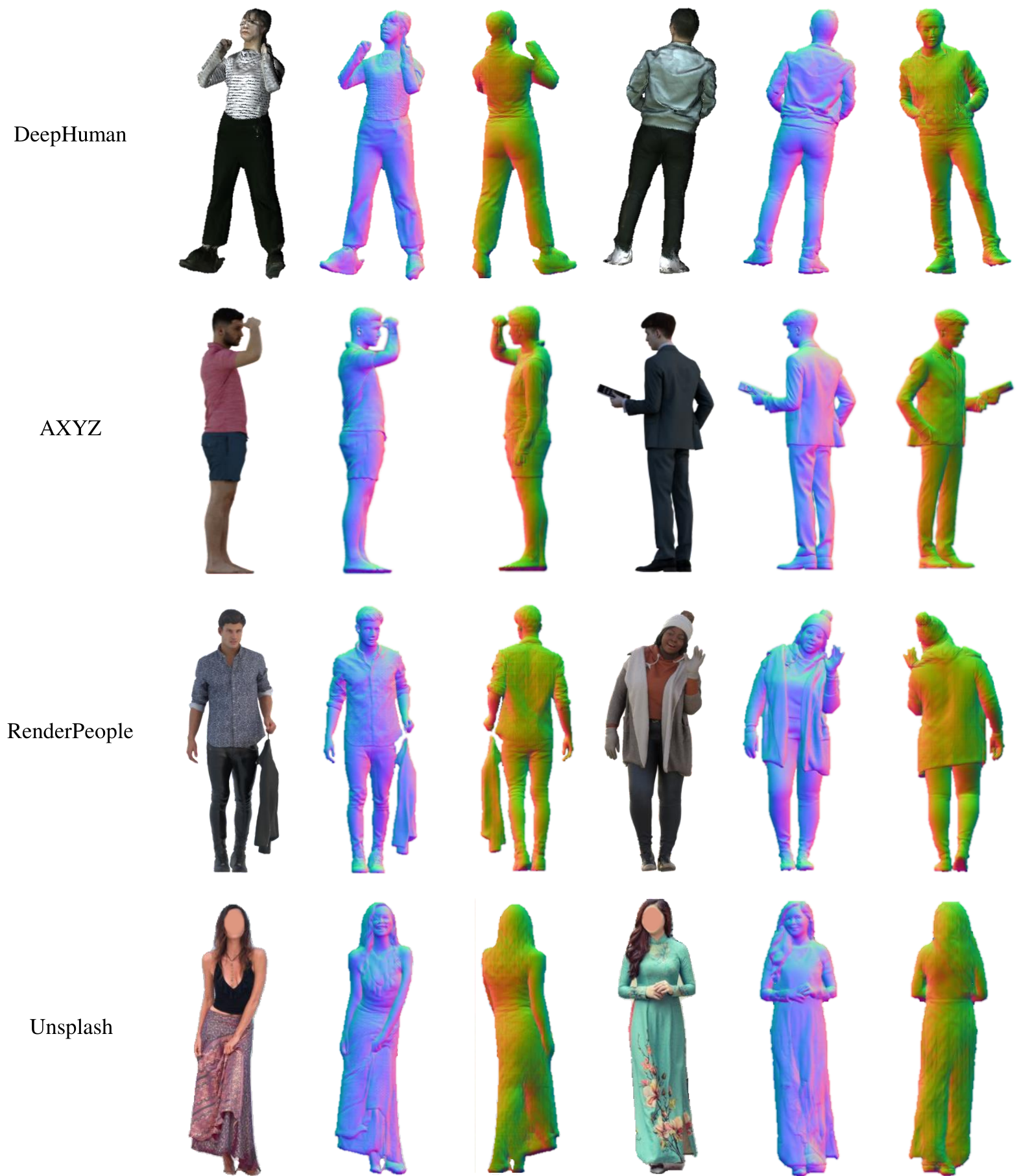}
\beforefigcaption
\vspace{+7 pt}
\caption{\textit{More normal refinement results from our approach on DeepHuman, AXYZ, RenderPeople and Unsplash datasets}, covering people with different camera views, poses and clothes. 
}
\afterfigcaption
\label{fig:supp_more_results}
\vspace{+20 pt}
\end{figure*}

\begin{figure*}[ptb]
\centering
\includegraphics[width=\textwidth]{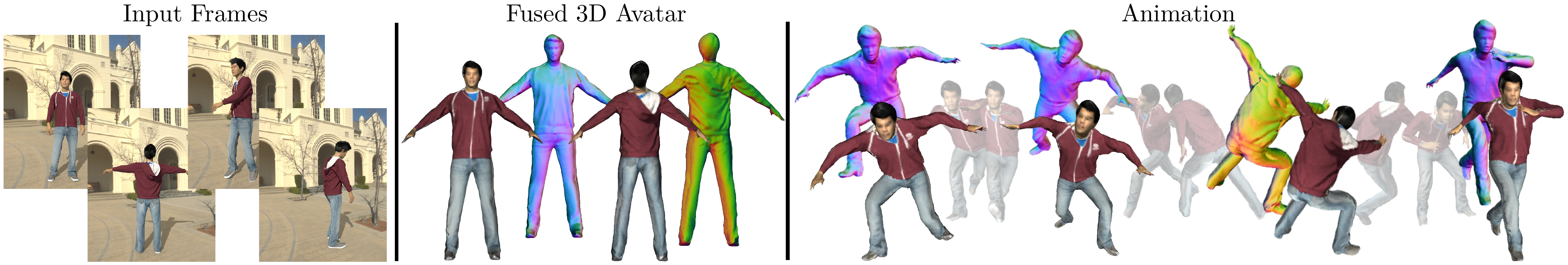}
\beforefigcaption
\vspace{+4pt}
\caption{\textit{An application of social telepresence}. Given an input video, our method can generate a fused 3D human avatar and further animate it with pre-defined Mixamo motions~\cite{mixamo}.
}
\afterfigcaption
\label{fig:animation}
\vspace{+13 pt}
\end{figure*}

\begin{figure*}[ptb]
\centering
\includegraphics[width=\linewidth]{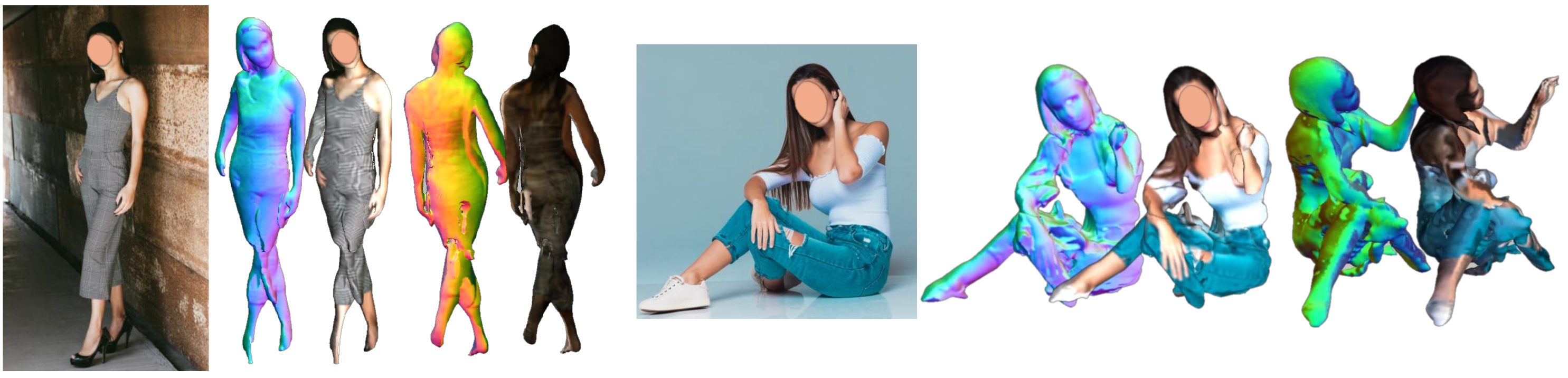}
\beforefigcaption
\vspace{+3pt}
\caption{\textit{Failure cases}. Reconstruction with strong directional lights and rare poses could be further improved.}
\afterfigcaption
\label{fig:failure_case}
\vspace{+8 pt}
\end{figure*}

\subsection{GAN Improves Back-Side Texture Estimation}

In our experiments, we found that GAN losses help to enhance the realism of back-side texture maps. In Fig.~\ref{fig:supp_texture_gan}, we demonstrate that the estimated textures with GAN training contain more plausible texture details and better lighting effect than those without GAN.

\subsection{Estimating Shaded Textures Preserves More Details}

Notably PIFu and ARCH all choose the albedo color space to predict. This requires the neural network to implicitly learn to compensate light/shading from the given shaded input images. However our current data scale seems insufficient to capture such a complicated space and might produce erroneous reconstructed textures. We believe our best strategy is to ``reconstruct the texture as similar/compatible as the input image``. We conduct ablation studies on learning different color spaces and the results are shown in Fig.~\ref{fig:supp_albedo_gan}. It can be observed that estimating back-side shaded textures, in comparison with albedos, preserves more details and overall generates similar and consistent color space to the input images.

\subsection{Inpainting Is a Needed Step for Avatar Reconstruction}

Compared with PIFu and ARCH, we propose to utilize the image space characteristics to better refine the reconstructed geometry. Our insight is that such features are naturally encoded in the image and there are already lots of powerful generative models in the literature which can solve similar tasks. However, one issue when trying to estimate the surface normals and textures lies in the missing surfaces cannot be ray traced from either the front or the back side, \eg, the occluded clothes by the arms in Fig.~\ref{fig:inpaint}. As a result, we believe such missing surfaces could be formulated and solved as an inpainting task. As shown in the figure, we are able to inpaint those missing surfaces (marked as gray regions) using the context. We could also observe that the final reconstructed avatar in the canonical space looks more complete and realistic than the one obtained by ARCH via implicit color field interpolation.


\subsection{More Normal Refinement Results}

We show more qualitative results on the testing sets (\eg, DeepHuman, AXYZ, RenderPeople and Unsplash) for people in different camera views, poses and clothes in Fig.~\ref{fig:supp_more_results}. It can be observe that our normal refinement results achieve high fidelity.

\subsection{User Study on Reconstruction Photorealism Shows Superior Quality of Our Method}

We set up a user study to further evaluate the photorealism of our method against other state of the art. In our study, we randomly pick \num{30} examples from our testing set (\ie, RenderPeople, AXYZ, Unsplash) and run the comparison methods~\cite{PIFuICCV19,huang2020arch,saito2020pifuhd} to obtain the results. For each example, we showed the input image and side-by-side rendered reconstruction results from two approaches in both front and back views to the participants. We numerate all pairs of approaches following a ``similarity judgment'' design~\cite{mantiuk2012comparison}. 
For each pair, the participants were asked to choose the more realistic result (``Which reconstructed avatar looks more real given the input image?''). They could choose either results or indicate that they were equally real.
To avoid biases and learning effects, we randomized the order of pairs as well as the position of the results while ensuring that for each participant the same number of techniques were shown on the left and the right sides.
To evaluate the results, we attribute the choice of one technique with $+1$ and the other with $-1$.
Averaging over all results in a quality score for each pair of techniques.
Using a t-test, we determine the probability of the drawn sample to come from a zero-mean distribution---zero-mean would indicate both techniques to be of equal quality.

We recruited 22 participants from universities and research institutes. Most participants are with medium to high experiences for computer vision and graphics. The results of the user study are summarized in Fig.~\ref{fig:user_study}. We conduct two sessions with one studying the normal reconstruction quality and the other one studying the texture reconstruction quality. We compute the p-value to indicates statistical significant difference between the methods according to a t-test (all significant results achieved at least p value $<.001$). All three pair-wise comparisons showed significant results:

For normal reconstruction, Ours vs. PIFuHD (mean $=0.39697$, std $=0.34625$, $t(21)=4.02$, $p < .001$), Ours vs. ARCH (mean $=0.83636$, std $=0.15535$, $t(21) = 25.25$, $p < .001$), and PIFuHD vs. ARCH (mean $=0.32121$, std $=0.27826$, $t(21)=5.41$, $p < .001$), with ours being significantly better than other state of the art and PIFuHD being significantly better than ARCH.

For texture reconstruction, Ours vs. PIFu (mean $=0.49091$, std $=0.21513$, $t(21)=10.70$, $p < .001$), Ours vs. ARCH (mean $=0.74545$, std $=0.21615$, $t(21)=16.18$, $p < .001$), and ARCH vs. PIFu (mean $=0.22424$, std $=0.21134$, $t(21)=4.98$, $p < .001$), with ours being significantly better than other state of the art and ARCH being significantly better than PIFu. 



\subsection{Applications}

When multi-view inputs are available (\eg, monocular videos), our method naturally supports canonical space normal and texture fusion to recover a photorealistic and animatable avatar thanks to the shared canonical space among different poses. For mesh vertices that are co-visible under multiple viewpoints we apply a simple yet effective normal-based linear blending scheme.
For each vertex, the normal/texture fusion weights \wrt one visible view is determined by the angle between the (unrefined)normal of that vertex and the camera direction. This is similar to Eq. (6) and (7) of the main paper. Here we further extend them to multiple views. Namely, one image that is facing towards the surface is weighted higher than another image of a large viewing angle.
As show in Fig.~\ref{fig:animation}, we fuse the normal and texture from multiple frames, and further generate an animation sequence.

\subsection{Failure cases}

As shown in Fig.~\ref{fig:failure_case}, there are some typical failure cases due to strong directional lighting and challenging poses. To tackle these issues we plan to add more lighting augmentation for the back-side texture estimation Pix2Pix module and also increase the size of our training scan set. For example, compared with the widely used image classification and detection datasets like ImageNet~\cite{deng2009imagenet,russakovsky2015imagenet} and COCO~\cite{lin2014microsoft}, our training dataset is relatively small consisting of only hundreds of 3D scans. Building a large-scale and high-quality clothed human mesh dataset with sufficient clothes types and human pose/shape variations is critical for pushing research works in image-based photorealistic avatar reconstruction.

\end{document}